\documentclass[lettersize,journal]{IEEEtran}
\usepackage{amsmath,amsfonts}
\usepackage{algorithmic}
\usepackage{algorithm}
\usepackage{array}
\usepackage[caption=false,font=normalsize,labelfont=sf,textfont=sf]{subfig}
\usepackage{booktabs}
\usepackage{textcomp}
\usepackage{stfloats}
\usepackage{url}
\usepackage{verbatim}
\usepackage{graphicx}
\usepackage{cite}
\usepackage{amssymb, pifont}
\usepackage{multirow}
\usepackage{booktabs}
\usepackage{graphicx}
\usepackage{color,xcolor}
\usepackage{hyperref}
\usepackage{threeparttable}
\usepackage{multirow}
\usepackage{amsmath}
\makeatletter
\renewcommand{\maketag@@@}[1]{\hbox{\m@th\normalsize\normalfont#1}}%
\makeatother

\newcommand{\etal}{\emph{et al.}}
\begin{document}
% Rethinking Image Dehazing via Laplace-based Frequency Enhancement and Hybrid Mamba-CNN Architecture
% Laplace-Mamba: When Laplace-based Frequency Enhancement Meets Hybrid Mamba-CNN Architecture for Image Dehazing
% Laplace Frequency-Enhanced Hybrid Mamba-CNN Architecture for Image Dehazing
%Laplace-Mamba: Laplace Frequency Prior Guided Hybrid Mamba-CNN for Image Dehazing
\title{Laplace-Mamba: Laplace Frequency Prior-Guided Mamba-CNN Fusion Network for Image Dehazing}

\author{Yongzhen Wang, Liangliang Chen, Bingwen Hu, Heng Liu, Xiao-Ping Zhang, \textit{Fellow}, \textit{IEEE}, and Mingqiang Wei, \textit{Senior Member}, \textit{IEEE}
        % , Yiping Chen, \textit{Senior Member}, \textit{IEEE}, Xiao-Ping Zhang, \textit{Fellow}, \textit{IEEE},
        % and Jonathan Li, \textit{Senior Member}, \textit{IEEE}% <-this % stops a space
\thanks{Yongzhen Wang, Liangliang Chen, Bingwen Hu, and Heng Liu are with the School of Computer Science and Technology, Anhui University of Technology, Ma’anshan 243032, China (e-mail: wangyz@ahut.edu.cn; chenll@ahut.edu.cn; hu\_bingwen@ahut.edu.cn; hengliu@ahut.edu.cn).}
\thanks{Xiao-Ping Zhang is with the Tsinghua Shenzhen International Graduate School, Tsinghua University, Shenzhen 518055, China (e-mail: xpzhang@ieee.org).}
\thanks{Mingqiang Wei is with the School of Computer Science and Technology, Nanjing University of Aeronautics and Astronautics, Nanjing 210016, China, and also with the College of Artificial Intelligence, Taiyuan University of Technology, Taiyuan 030024, China (e-mail: mingqiang.wei@gmail.com).}
% \thanks{\textit{(Corresponding author: Xuefeng Yan.)} }

% \thanks{Y. Chen is with the Fujian Key Laboratory of Sensing and Computing for Smart Cities, School of Informatics, Xiamen University, Xiamen 361005, China (chenyiping@xmu.edu.cn). }

% \thanks{J. Li is with the Department of Geography and Environmental Management and Department of Systems Design Engineering, University of Waterloo, Waterloo, Canada (e-mail: junli@uwaterloo.ca). }
% \thanks{W. Wang is with the School of Science and Technology, Hong Kong Metropolitan University, Hong Kong, China (e-mail: wmwang@hkmu.edu.hk)}
% \thanks{H. Xie is with the Department of Computing and Decision Sciences, Lingnan University, New Territories, Hong Kong, China (hrxie2@gmail.com).} 
% \thanks{Y. Chen is with the Fujian Key Laboratory of Sensing and Computing for Smart Cities, School of Informatics, Xiamen University, Xiamen, China (e-mail: chenyiping@xmu.edu.cn).}

% \thanks{X.-P. Zhang is with the Department of Electrical, Computer and Biomedical Engineering, Ryerson University, Toronto, Canada (e-mail: xzhang@ee.ryerson.ca).} 

% \thanks{J. Li is with the Department of Geography and Environmental Management University of Waterloo, Waterloo, Canada (e-mail: junli@uwaterloo.ca). }
% \thanks{\textit{(Corresponding author: Xuefeng Yan.)} }
}

\markboth{Journal of \LaTeX\ Class Files,~Vol.~14, No.~8, August~2024}%
{Shell \MakeLowercase{\textit{et al}}: A Sample Article Using IEEEtran.cls for IEEE Journals}


\maketitle
\begin{abstract}
Recent progress in image restoration has underscored Spatial State Models (SSMs) as powerful tools for modeling long-range dependencies, owing to their appealing linear complexity and computational efficiency. However, SSM-based approaches exhibit limitations in reconstructing localized structures and tend to be less effective when handling high-dimensional data, frequently resulting in suboptimal recovery of fine image features. To tackle these challenges, we introduce Laplace-Mamba, a novel framework that integrates Laplace frequency prior with a hybrid Mamba-CNN architecture for efficient image dehazing. Leveraging the Laplace decomposition, the image is disentangled into low-frequency components capturing global texture and high-frequency components representing edges and fine details. This decomposition enables specialized processing via dual parallel pathways: the low-frequency branch employs SSMs for global context modeling, while the high-frequency branch utilizes CNNs to refine local structural details, effectively addressing diverse haze scenarios. Notably, the Laplace transformation facilitates information-preserving downsampling of low-frequency components in accordance with the Nyquist theory, thereby significantly improving computational efficiency. Extensive evaluations across multiple benchmarks demonstrate that our method outperforms state-of-the-art approaches in both restoration quality and efficiency. The source code and pretrained models are available at \textcolor{magenta}{\href{https://github.com/yz-wang/Laplace-Mamba}{https://github.com/yz-wang/Laplace-Mamba}}.
\end{abstract}


%-------------------------------------------------------------------------
\section{Introduction}
\IEEEPARstart{A}{s} a crucial preprocessing component for robust visual systems, single image dehazing endeavors to restore latent clear scenes from degraded observations, a task of particular significance for vision applications in adverse weather conditions\cite{he2010single}. This inherently ill-posed inverse problem arises from the irreversible information loss during atmospheric degradation, wherein only hazy inputs are accessible without corresponding clear references. To tackle this challenge, numerous traditional techniques have been introduced\cite{he2010single}\cite{cai2016dehazenet}\cite{meng2013efficient} \cite{zhu2015fast}\cite{berman2016non}, encompassing methods rooted in atmospheric scattering models and hand-crafted dehazing priors. While these methods offer certain advantages, they often exhibit limited generalization capability and fail to effectively address complex haze scenarios.

\begin{figure}[t] 
    \centering
    \includegraphics[width=1.0\linewidth]{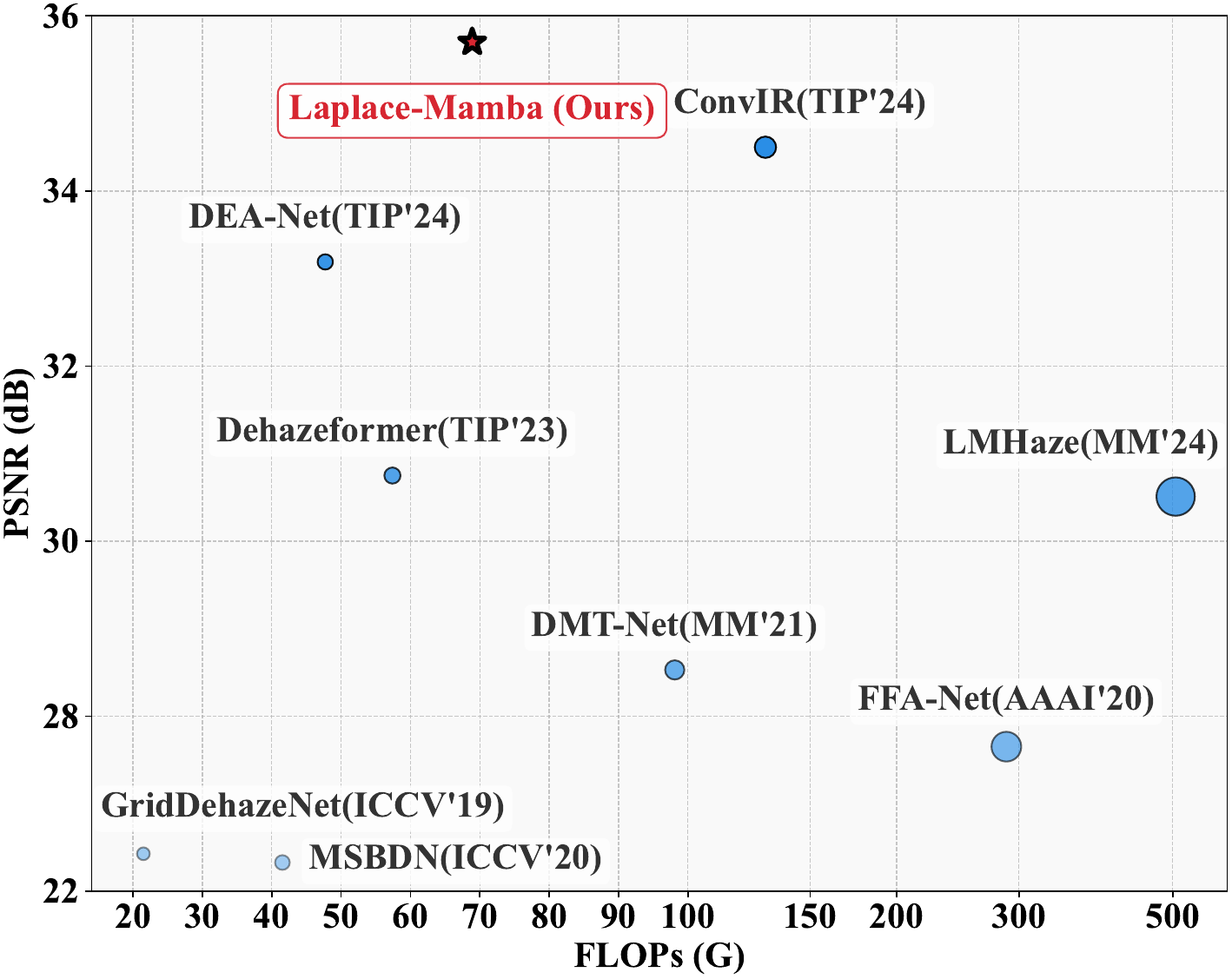}
	\caption{Comparison of computational cost and PSNR for each model on the Haze4K dataset \cite{song2021haze4k}. Notably, our Laplace-Mamba achieves efficient and high-quality feature restoration by integrating Laplace frequency prior with Mamba-CNN hybrid architecture.}
 	\label{fig:fig1} 
 \end{figure}

In recent years, the remarkable success of deep learning in computer vision has revolutionized image dehazing through the development of various neural network-based methodologies. Early efforts predominantly employ Convolutional Neural Network (CNN) architectures \cite{cai2016dehazenet}, \cite{zhang2018dcpdn}, \cite{li2018dehaze_cgan}, \cite{engin2018cycle_dehaze}, encompassing encoder-decoder frameworks, multi-stage networks, and dual-stream architectures, among others. While these approaches achieved notable progress, they are intrinsically constrained by limited receptive fields and the inability to effectively model global contextual dependencies, which undermines their overall dehazing capability. To overcome these limitations, growing attention has been directed toward Generative Adversarial Networks (GANs) for image dehazing. GAN-based solutions, such as conditional GANs \cite{li2018dehaze_cgan}, cycle-consistent GANs \cite{engin2018cycle_dehaze}, and attention-enhanced GANs \cite{wang2019aagan}, have been proposed to enhance the perceptual quality of dehazed outputs. Despite yielding promising results, these models often face challenges such as training instability and difficulty in preserving fine-grained structural details, thereby limiting their robustness in complex and dynamically varying hazy scenes. Drawing inspiration from the success of Transformers in modeling global dependencies and capturing long-range feature interactions, researchers have increasingly adapted transformer-based architectures\cite{valanarasu2021transweather} for image dehazing. These approaches offer several advantages over conventional learning-based methods: they exhibit superior training stability relative to GAN-based models and benefit from larger receptive fields compared to CNN-based counterparts. This synergy enhances their capacity to model long-range dependencies, a critical factor for effective image dehazing. However, the self-attention mechanism inherent to Transformers introduces significant computational overhead due to its quadratic complexity, presenting a practical limitation in real-world applications.
% To mitigate this, some methods have adopted channel-wise attention mechanisms\cite{wang2020ecanet}, but these often fail to fully exploit spatial information, compromising dehazing performance. While window-based self-attention mechanisms\cite{liang2021swinir} have been proposed to address this issue, their patch-based segmentation strategies still struggle to comprehensively explore inter-patch information. 

Recently, Spatial State Models (SSMs)\cite{gu2023mamba} have garnered increasing attention in computer vision for their capacity to model intricate long-range dependencies while maintaining linear computational complexity. Despite their efficiency, the application of Mamba-based models to image processing tasks often results in local pixel forgetting \cite{vim}, a phenomenon that undermines the preservation of fine-grained structures and consequently degrades the overall quality of image restoration. Therefore, decoupling the processing of global and local features enables more effective image restoration. Moreover, most existing approaches tend to overlook the potential of frequency domain information, which serves as a valuable complement to spatial domain cues. Integrating such information can significantly enrich image representations, thereby contributing to more accurate and robust image recovery.

% To address this issue, MambaIR \cite{guo2025mambair} introduces a four-directional scanning strategy to alleviate local pixel forgetting, which effectively mitigates the problem of localized image blurring. However, this approach also results in significant computational inefficiency, leading to extended image inference times.

Inspired by these observations, we present Laplace-Mamba, a novel image dehazing framework that seamlessly integrates Laplace frequency decomposition with a hybrid Mamba-CNN architecture for efficient and effective restoration. Laplace-Mamba decomposes image features into high- and low-frequency components and processes them independently, thereby overcoming the limitations of conventional approaches in jointly modeling local and global information. By separately processing the local details (high-frequency components) and global structures (low-frequency components), the framework achieves superior restoration fidelity, particularly in enhancing fine-grained details. Furthermore, our approach capitalizes on the Laplace transform's intrinsic property of preserving information during low-frequency feature downsampling, leading to significant computational efficiency gains. The proposed framework specifically exploits spectral characteristics unique to hazy images: high-frequency bands primarily encode local edge information and contours, whereas low-frequency domains encapsulate global structural patterns. Unlike prior Mamba-based models, Laplace-Mamba processes these components separately: the Low-frequency Structure Restoration Block (LSRB) is employed to reconstruct global structures, while the High-frequency Detail Enhancement Block (HDEB) is designed to refine local textures. As illustrated in Fig. \ref{fig:fig1}, our Laplace-Mamba not only achieves superior performance but also demonstrates a significant reduction in computational overhead compared to existing approaches. In summary, the key contributions of this work are as follows:
\begin{itemize}
    \item We propose Laplace-Mamba, a novel image dehazing framework that integrates Laplace frequency prior with a hybrid Mamba-CNN architecture, effectively enhancing dehazing performance while preserving high computational efficiency.
    \item We develop a Multi-Domain Fusion Module that enhances global representations in the low-frequency branch by integrating complementary features from multiple domains, effectively preserving structural coherence and improving overall dehazing performance.
    \item We propose a novel Frequency-Domain ‌Collaborative Module tailored to the distinct characteristics of frequency components: the Low-frequency Structure Restoration Block reconstructs global structures, while the High-frequency Detail Enhancement Block refines local textures, leading to significantly improved detail recovery in challenging hazy conditions.
    % \item Extensive experiments demonstrate that Laplace-Mamba achieves state-of-the-art performance in image dehazing tasks, balancing effectiveness and efficiency.
\end{itemize}

The remainder of this paper is organized as follows: Section II provides a comprehensive review of related work, encompassing prior-based methods, learning-based approaches, and Spatial State Models. Section III introduces the proposed Laplace-Mamba framework in detail. Section IV presents extensive experimental results and performance analysis. Finally, Section V concludes the paper with a summary of the findings and contributions.

\section{Related Work}
%In this section, we will briefly review the development of image dehazing techniques and discuss the application of spatial state models.

\subsection{Prior-based Methods}
% Early image dehazing methods relied on handcrafted priors and a priori knowledge to address this ill-posed problem. These methods typically used assumptions like dark channel priors and atmospheric scattering models to estimate scene transmission and atmospheric light. For example, the Dark Channel Prior (DCP) \cite{he2010single} assumes that one color channel in haze-free images has very low intensity values, effective in many outdoor scenes. The Color Attenuation Prior (CAP) \cite{zhu2015fast} uses a linear model of image brightness and saturation to estimate depth, enabling faster inference. The Non-local Prior \cite{berman2016non} leverages pixel alignment in RGB space to improve transmission estimation through color clustering. The Boundary Constrained Prior \cite{meng2013efficient} refines transmission maps with edge-aware constraints. The Multi-scale Fusion Prior \cite{berman2018single} combines haze estimates across multiple scales for smoother results. However, these methods' reliance on simplified physical assumptions limits their applicability in complex real-world scenes, often resulting in suboptimal dehazing performance and computational inefficiency.

Early image dehazing approaches predominantly leveraged handcrafted priors and domain-specific knowledge to tackle the inherently ill-posed nature of this problem. These methods typically utilize predefined assumptions regarding scene transmission and atmospheric light to restore clear images.
One of the most influential techniques, the Dark Channel Prior (DCP) \cite{he2010single}, operates on the assumption that at least one color channel in haze-free images exhibits very low intensity values, making it highly effective for outdoor scenes. Extending this concept, the Color Attenuation Prior (CAP) \cite{zhu2015fast} introduced a linear model linking image brightness and saturation to scene depth estimation, achieving faster inference without significantly compromising accuracy.
The Non-local Prior \cite{berman2016non} further advanced the field by leveraging pixel alignment in the RGB space, effectively improving transmission estimation through color clustering. Meanwhile, the Boundary Constrained Prior (BCP) \cite{meng2013efficient} refined transmission maps using edge-aware constraints, enhancing dehazing quality around object boundaries. Multi-scale Fusion Prior (MFP) \cite{berman2018single} proposed a multi-scale strategy integrating haze estimates across various resolutions, producing smoother and more consistent dehazing results.

Despite their initial success, these prior-based methods are fundamentally constrained by their reliance on simplified physical models and handcrafted assumptions, limiting their generalizability and robustness in complex real-world scenarios. As a result, their performance often degrades in the presence of diverse haze conditions, non-uniform illumination, and dense atmospheric scattering, highlighting the need for more sophisticated data-driven solutions.

\subsection{Learning-based Methods}

The rise of deep learning has revolutionized image dehazing, with numerous CNN-based methods emerging. Early models primarily estimated transmission maps via neural networks and reconstructed images using the atmospheric scattering model. For instance, Zamir \etal \cite{zamir2021multi} introduced an inter-stage feature fusion network for capturing multi-scale features, while Cui \etal proposed the Omni-Kernel Network (OKN) \cite{cui2024omni}, leveraging bi-domain processing and large-kernel convolutions for better contextual understanding. However, these CNN-based methods remain limited by their inability to capture long-range spatial dependencies due to the localized nature of convolution operations.
Generative Adversarial Networks (GANs) have also been applied to image dehazing, excelling in texture restoration and perceptual quality. Notable approaches include AOD-GAN \cite{li2018single}, which integrates adversarial learning with the atmospheric scattering model, Cycle-Dehaze \cite{engin2018cycle}, which uses unpaired data and cycle-consistent learning, and a multi-scale network for nighttime dehazing proposed by Zhang \etal \cite{zhang2020nighttime}. Despite their strengths, GAN-based methods often suffer from training instability and mode collapse, especially under diverse haze conditions. In addition, Vision Transformers (ViTs) have advanced image dehazing by modeling global dependencies. Zhao \etal \cite{zhao2022local} first applied ViTs to dehazing with a local-global architecture, but their quadratic self-attention complexity remains a challenge for high-resolution images. To address this, Uformer \cite{wang2022uformer}, SwinIR \cite{liang2021swinir}, and U2former \cite{chen2022u2former} use window-based self-attention, though at the cost of fragmented spatial modeling. Alternatives like Restormer \cite{zamir2022restormer} and MRLPFNet \cite{li2022mrlpfnet} adopt channel-wise attention for efficiency, sacrificing spatial detail. FFTformer \cite{wang2023fftformer} offers a balance by leveraging frequency-domain properties for scalable attention.

Despite the impressive performance of these learning-based methods, they still encounter limitations in balancing local detail preservation and global context modeling, particularly in complex, dense, or non-uniform haze conditions. Our proposed Laplace-Mamba method addresses these issues by integrating frequency-domain analysis with a hybrid model, effectively capturing both local and global features in a computationally efficient manner.

\subsection{Spatial State Models}
%Recent breakthroughs in spatial state models (SSMs) have introduced a paradigm shift for long-range dependency modeling with linear computational complexity. Structured state space models, such as S4 \cite{gu2021efficiently}(the first SSM for sequence modeling) and S5\cite{smith2022simplified} (featuring parallelizable recurrent computation), laid the groundwork for efficient sequence processing. The selective scanning mechanism in S6\cite{gu2023mamba} further enhanced adaptability, positioning SSMs as a competitive alternative to transformers. Vision Mamba\cite{zhu2024vision} extended SSMs to vision tasks as a general-purpose backbone, while MambaIR\cite{guo2025mambair} introduced a four-directional scanning strategy to mitigate local information loss. Wave-Mamba\cite{wang2024wave} integrated wavelet priors to enhance multi-scale feature extraction. Despite these advances, Gao \etal \cite{gao2024learning} identified persistent issues of local pixel forgetting in SSMs, proposing hybrid SSM-CNN architectures for improved dehazing efficacy.

% In the paper, we train the Transformer dehazing network with the guidance of high-level semantic knowledge, facilitating to generate high-quality perceptual results. Existing dehazing approaches focus on solving low-level problems and high-level problem separately, which bridges the gap between high-level and low-level tasks obviously. By leveraging high-level vision semantic information, our method not only contributes to recovering object structures, details and colors from dehazed images, but also avoids the gap effectively. 

Spatial state models (SSMs) have recently emerged as powerful tools for modeling long-range dependencies with linear computational complexity. The development of SSMs began with S4 \cite{gu2021efficiently}, the first model designed specifically for sequence modeling, which introduced efficient state-space processing. This was followed by S5 \cite{smith2022simplified}, which improved processing speed through parallelizable recurrent computation. Further advancements came with S6 \cite{gu2023mamba}, which incorporated a selective scanning mechanism to enhance adaptability, positioning SSMs as viable alternatives to transformers in sequence tasks.
Following their success in sequence modeling, SSMs have been successfully extended to vision applications. Vision Mamba \cite{zhu2024vision} showcased their potential as a versatile backbone, while MambaIR \cite{guo2025mambair} addressed local information loss through a four-directional scanning strategy. To improve multi-scale feature extraction, Wave-Mamba \cite{wang2024wave} integrated wavelet priors, offering a more comprehensive representation of image details. Despite these achievements, SSMs still struggle with capturing fine-grained local structures, particularly in high-resolution image restoration. To overcome this, Gao \etal \cite{gao2024learning} proposed hybrid SSM-CNN architectures, combining the global modeling capabilities of SSMs with the local feature extraction strengths of CNNs. 

Inspired by the hybrid SSM-CNN architecture, our proposed Laplace-Mamba further extends this concept by combining the strengths of SSMs and CNNs in a hybrid framework while introducing frequency-domain analysis. Specifically, we leverage Laplace spectral decomposition to disentangle image features into low-frequency global context and high-frequency local details, processed independently to ensure more efficient and accurate haze removal.

\begin{figure*}[htbp] 
    \centering
    \includegraphics[width=0.9\linewidth]{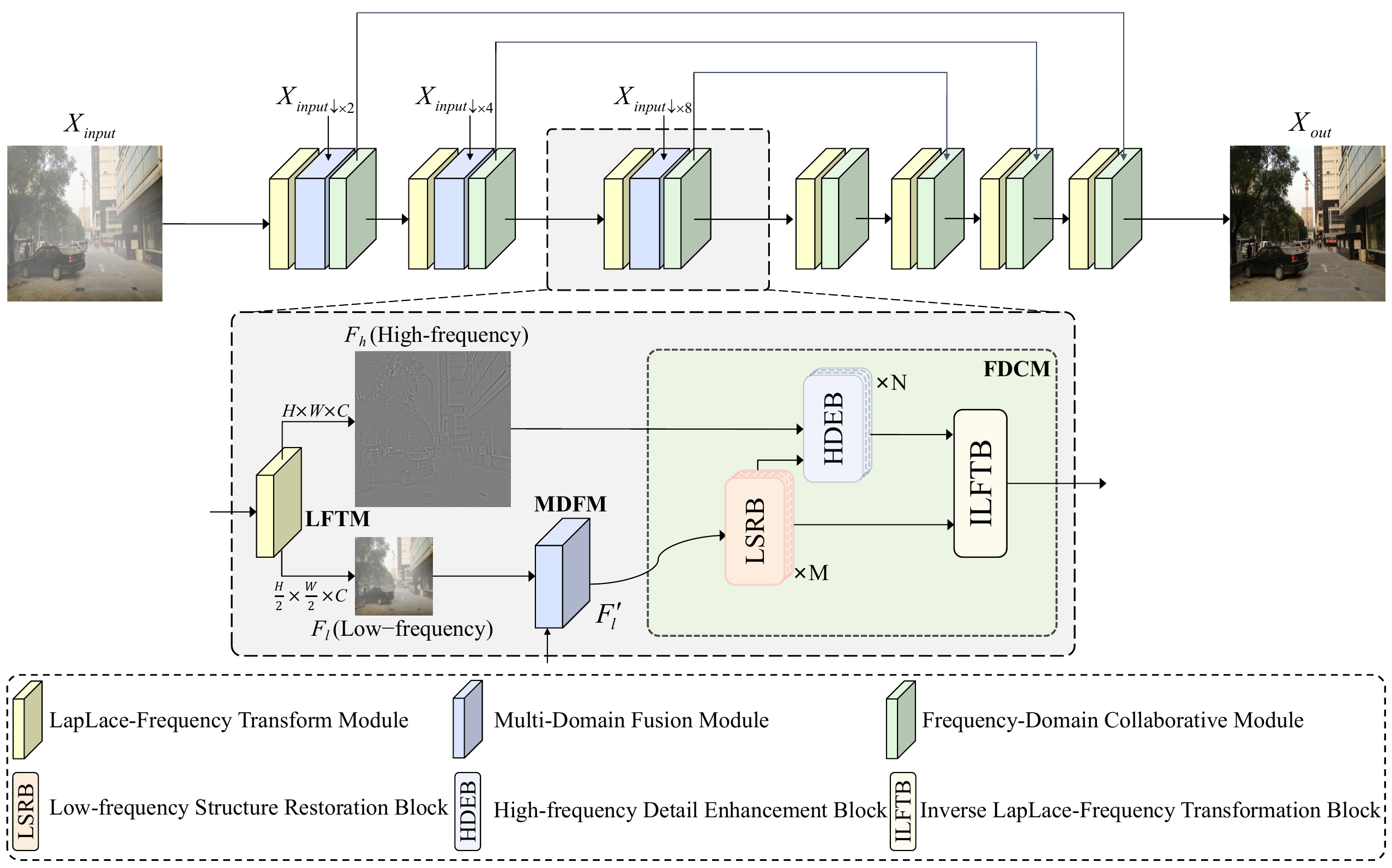}
	\caption{
        Overview of Laplace-Mamba. Laplace-Mamba is composed of three primary components: (1) a Laplace-Frequency Transform Module(LFTM) that performs an information-preserving decomposition into high- and low-frequency components; (2) a Multi-Domain Fusion Module(MDFM) designed to enhance the representational richness of the low-frequency features by integrating spatial and frequency-domain information; and (3) a Frequency-Domain Collaborative Module(FDCM), which leverages the complementary strengths of Mamba and CNN to separately model global low-frequency structures and high-frequency local details for enhanced restoration fidelity.
 	}
 	\label{fig:fig2}
 \end{figure*}

\section{Laplace-Mamba}
Laplace-Mamba seamlessly integrates Laplace frequency prior knowledge with the complementary strengths of Mamba and CNN architectures, enabling effective capture of both fine-grained local details and broad contextual information across high/low-frequency bands. By producing low-frequency components at half the resolution of the original input, the framework substantially reduces the computational burden of the restoration process while preserving the fidelity of frequency decomposition and reconstruction. This design sustains restoration quality while reducing computational requirements, particularly within the Mamba-based modules. To enrich the global representation of low-frequency features, a novel Multi-Domain Fusion Module is introduced, facilitating more robust modeling of long-range dependencies. Furthermore, Laplace-Mamba leverages the reconstructed low-frequency image features to guide the enhancement of high-frequency components, enabling effective reconstruction of global-local features across different frequency bands and significantly improving overall dehazing performance.
% To optimize the reconstruction, the framework performs local feature extraction through the recovered low-frequency image while enhancing the recovery of the high-frequency components using the attention mechanism to achieve a comprehensive detail reconstruction across all frequency domains.

% In this section, we first present an overview of the proposed Laplace-Mamba framework. We then provide a detailed description of the network architecture, encompassing the Laplace-Frequency Transform Module and the Multi-Domain Fusion Module. Subsequently, we elaborate on the Low-Frequency Structure Restoration Block and the High-Frequency Detail Enhancement Module, highlighting their respective roles and operational mechanisms within the Laplace-Mamba pipeline. Lastly, we introduce the loss functions used to train Laplace-Mamba.

\subsection{Overview}
The overall architecture of Laplace-Mamba, depicted in Fig. \ref{fig:fig2}, follows a U-Net-inspired design comprising three primary components: the Laplace-Frequency Transform Module (LFTM), the Multi-Domain Fusion Module (MDFM), and the Frequency-Domain ‌Collaborative Module (FDCM). Given an input hazy image $x_{input}$, Laplace-Mamba initially performs Laplace frequency decomposition through the LFTM, effectively disentangling the image into its low-frequency ($F_l$) and high-frequency ($F_h$) components. The low-frequency component $F_l$ is then fused with spatial features $F_s$ through the MDFM, enhancing its global representation to produce ($F_{l}'$) that captures more robust global context. Subsequently, this refined feature is processed by Low-frequency Structure Restoration Block within the FDCM, yielding an optimized global representation $F_l^*$. Simultaneously, the high-frequency component $F_h$ is input to the High-frequency Detail Enhancement Block, where it is guided by $F_l^*$ to facilitate context-aware enhancement, resulting in an improved high-frequency output $F_h^*$. Finally, an inverse Laplace transform is applied to $F_l^*$ and $F_h^*$, fusing them into the final dehazed image $x_{out}$. This architecture effectively combines multi-scale frequency restoration with cross-domain feature interaction to achieve high-fidelity dehazing.

\subsection{Laplace-Frequency Transform Module}
The Laplace-Frequency Transform Module (LFTM) leverages the Laplacian pyramid to decompose the input image into continuous frequency components, yielding a high-frequency component and a low-frequency counterpart. It is worth emphasizing that this decomposition process is inherently lossless, ensuring the faithful preservation of image content throughout the entire frequency separation and reconstruction. Moreover, as the low-frequency component undergoes a downsampling operation during decomposition, the subsequent processing stages benefit from reduced computational complexity without sacrificing representational integrity. The high-frequency component retains rapidly varying features such as textures, edges, and contours, while the low-frequency component encapsulates slowly varying structural cues, including background layout, color distribution, and illumination. Therefore, we posit that decoupled processing of high- and low-frequency components facilitates more precise and complementary restoration, thus enhancing overall image quality.

\begin{figure}[t] 
    \centering
    \includegraphics[width=\linewidth]{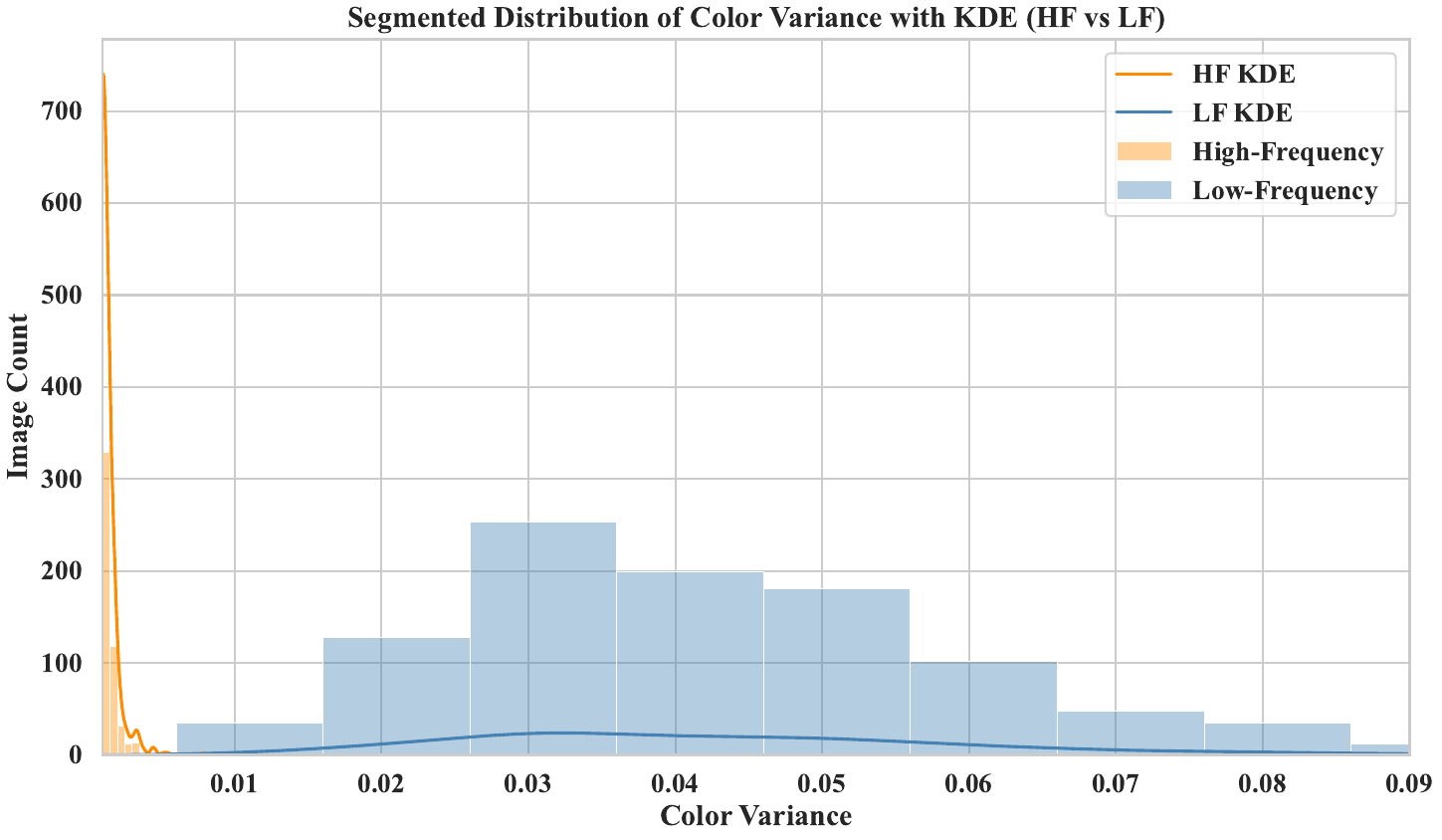}
	\caption{
        Color variance distribution across 1,000 image samples for high- and corresponding low-frequency components. Orange bars denote high-frequency components, while blue bars represent their low-frequency counterparts.}
 	\label{fig:fig3}
 \end{figure}

To substantiate our hypothesis that low-frequency components are more amenable to global modeling while high-frequency components are better suited for local optimization, we conduct a comprehensive statistical analysis of color variance across 1,000 image samples for both frequency domains. In this context, color variance serves as an indicator of chromatic complexity; higher variance denotes the presence of intricate textures and fine-grained details. In contrast, lower variance implies smoother, more homogeneous regions. As illustrated in Fig. \ref{fig:fig3}, the distribution of color variance in low-frequency components (depicted in blue) is consistently lower than that in high-frequency components (shown in orange). This contrast highlights the structural coherence of low-frequency information and the textural richness of high-frequency details. Motivated by this empirical observation, we design frequency-specific restoration modules that independently target each domain, enabling a more principled and effective approach to image dehazing.

\subsection{Multi-Domain Fusion Module}
Following Laplace frequency decomposition, global contextual information in low-frequency components may be diminished. Although U-Net-style skip connections help preserve spatial details, naive fusion strategies such as direct summation or concatenation often lead to the accumulation of redundant grayscale values from the encoder, thereby degrading restoration quality. To mitigate this, we incorporate an attention mechanism that dynamically regulates feature transmission, suppressing unnecessary grayscale propagation. Inspired by attention-based fusion techniques \cite{qin2020feature, zheng2023frequency}, we introduce a novel Multi-Domain Fusion Module (MDFM) that adaptively integrates multi-scale spatial features ($F_s$) with frequency-domain representations ($F_l$).

% Following Laplace frequency decomposition, a portion of the global contextual information contained in the low-frequency components may be diminished. To mitigate this, a common solution employed in U-Net architectures is the use of skip connections to supplement spatial information across corresponding encoder-decoder stages. However, naive fusion methods such as direct summation or concatenation can lead to the accumulation of grayscale values from the encoder to the decoder, ultimately degrading the restoration quality. To address this issue, we integrate an attention mechanism within the fusion module to dynamically modulate feature transmission, thereby attenuating redundant grayscale propagation. This approach is motivated by previous explorations into attention-enhanced spatial feature fusion, such as those proposed by Qin \etal~\cite{qin2020feature} and Zheng \etal~\cite{zheng2023frequency}, which primarily focused on single-scale enhancement. To further enhance spatial-frequency feature integration, we propose a Multi-Domain Fusion Module (MDFM) that effectively combines multi-scale spatial features with their corresponding frequency-domain representations. MDFM employs learnable adaptive weights to guide the feature fusion process. This adaptive fusion mechanism allows the model to more effectively emphasize informative features and suppress irrelevant noise, particularly improving the removal of shallow haze and enhancing the overall restoration quality across complex degradation scenarios.

\begin{figure}[t] 
    \centering
    \includegraphics[width=1.0\linewidth]{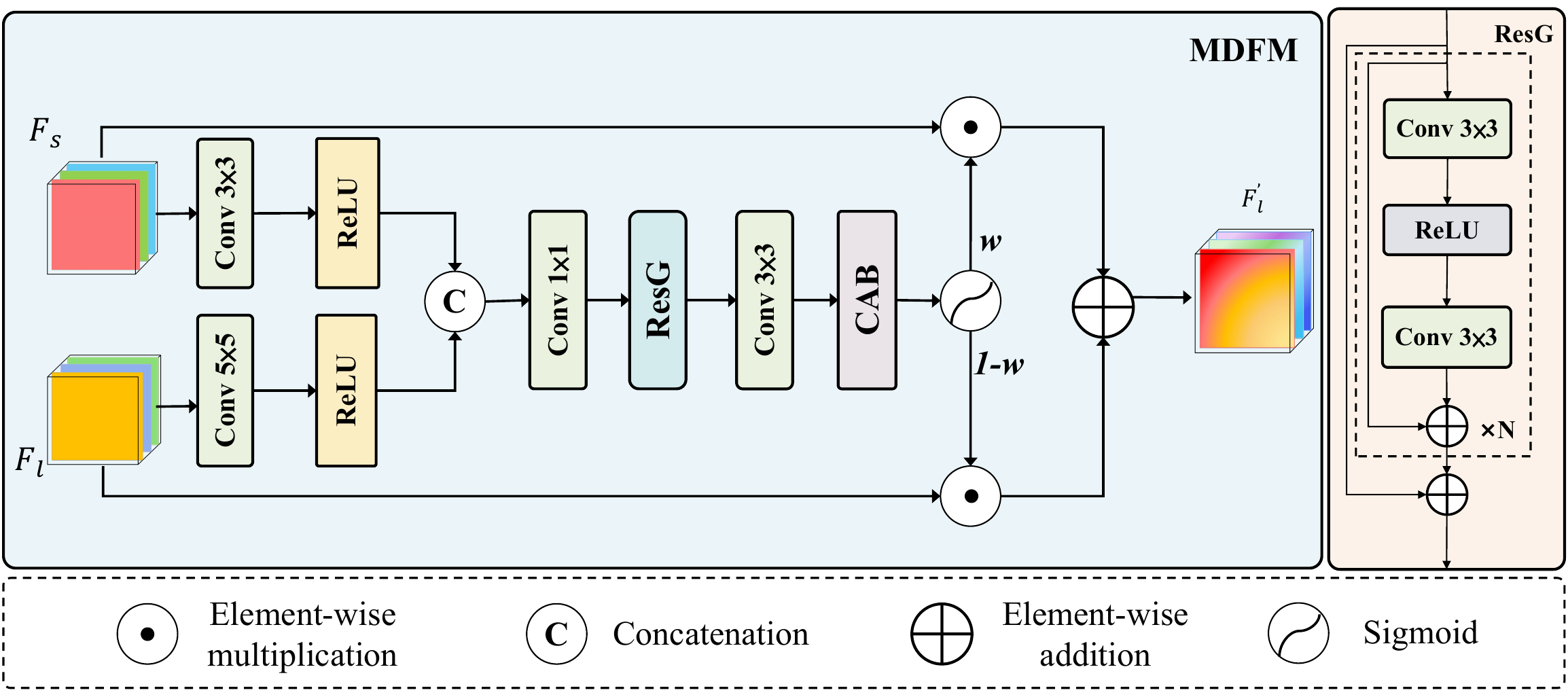}
	\caption{Architecture of the Multi-Domain Fusion Module(MDFM). 
 	}
 	\label{fig:fig10}
 \end{figure}

As illustrated in Fig. \ref{fig:fig10} , we aim to fuse two types of feature representations: spatial-domain features $F_s$ and frequency-domain features $F_l$, where $F_s, F_l \in \mathbb{R}^{H/2 \times W/2 \times C}$. The fusion process is carefully designed to leverage the complementary characteristics of both domains. First, each feature map undergoes a ReLU activation followed by convolutional filtering to extract salient features. Specifically, a 3$\times$3 convolution is applied to the spatial features $F_s$, while a 5$\times$5 convolution is applied to the frequency features $F_l$, resulting in intermediate representations $Y^1$ and $Y^2$:
\begin{align}
Y^1 &= \mathcal{A}(\textit{Conv}_3(F_s)), \\
Y^2 &= \mathcal{A}(\textit{Conv}_5(F_l)),
\end{align}
where $\mathcal{A}(\cdot)$ denotes the ReLU activation function and $\textit{Conv}_k(\cdot)$ represents a convolution operation with kernel size $k$$\times$$k$. The outputs $Y^1$ and $Y^2$ are then concatenated along the channel dimension and passed through a ResGroupNet module (3$\times$3 convolution combined with activation function) to generate the fusion weight map $W$:
\begin{align}
W = \mathcal{CA}\Big(\mathit{Conv}_1\big(\mathit{ResG}(\mathit{Conv}_1(\mathit{Concat}(Y^1, Y^2)))\big)\Big),
\end{align}
where $\mathcal{C}(\cdot)$ denotes channel-wise concatenation, $\text{ResG}(\cdot)$ refers to a residual block (see Fig.~\ref{fig:fig10} right), and $\mathcal{CA}(\cdot)$ is a channel attention mechanism that adaptively calibrates channel-wise responses. Finally, the fused output $F'_{\mathit{l}}$ is computed via a weighted element-wise combination of the spatial and frequency feature maps $X_s$ and $X_f$, respectively:
\begin{align}
F'_{\mathit{l}} = X_f \odot W + X_s \odot (1 - W),
\end{align}
where $\odot$ denotes element-wise multiplication. The resulting output $F'_{\mathit{l}}$ is subsequently used as input to the low-frequency restoration module, enhancing the reconstruction of global structural information.

% First, both feature maps are passed through the ReLU activation function, followed by convolutional layers with kernel sizes of 5 and 3 to extract key features, yielding $Y^1$ and $Y^2$. Next, $Y^1$ and $Y^2$ are concatenated and processed through ResGroupNet to obtain W. The ResGroupNet module consists of a $1\times1$ convolution at the front and a $3\times3$ convolution at the back, formulated as follows:
% \begin{align}
%     &Y^1,Y^2 = \mathcal{A}(Con_3(F_{s})),\mathcal{A}(Con_5(F_{l})),\\
%     &W = \mathcal{CA}(Conv(ResG(Conv(\mathcal{C}(Y_1,Y_2))),
% \end{align}
% where $C$ denotes a simple connection operation, $\mathcal{A}$ denotes a ReLU function, ResG denotes the simple residual block specific structure shown in Fig. \ref{fig:fig2}~(right), and Convk denotes $k\times k$ convolution, and $\mathcal{CA}$ denotes the channel attention mechanism.

% After obtaining the fusion weights $w$, the final output $x_{out}$ is obtained by multiplying $w$ and $1-w$ to $x_{f}$ and $x_{s}$ by the product of the elements, the formula is specified as follows:
% \begin{align}
%     &X_{out} = X_{f}\odot W+X_{s}\odot(1-W),
% \end{align}
% where $\odot$ indicates element-wise product. The output obtained is used as input to the low-frequency repair module.

\subsection{Frequency-Domain ‌Collaborative Module}
To effectively exploit the high- and low-frequency components extracted via Laplace decomposition, we propose a novel Frequency-Domain ‌Collaborative Module (FDCM). This module leverages the complementary strengths of Mamba and CNN architectures, enabling specialized processing of distinct frequency bands. By assigning Mamba to model global dependencies in low-frequency information and CNNs to refine local textures in high-frequency features, FDCM enhances the restoration quality through frequency-aware cooperative learning.

\begin{figure}[t] 
    \centering
    \includegraphics[width=0.65\linewidth]{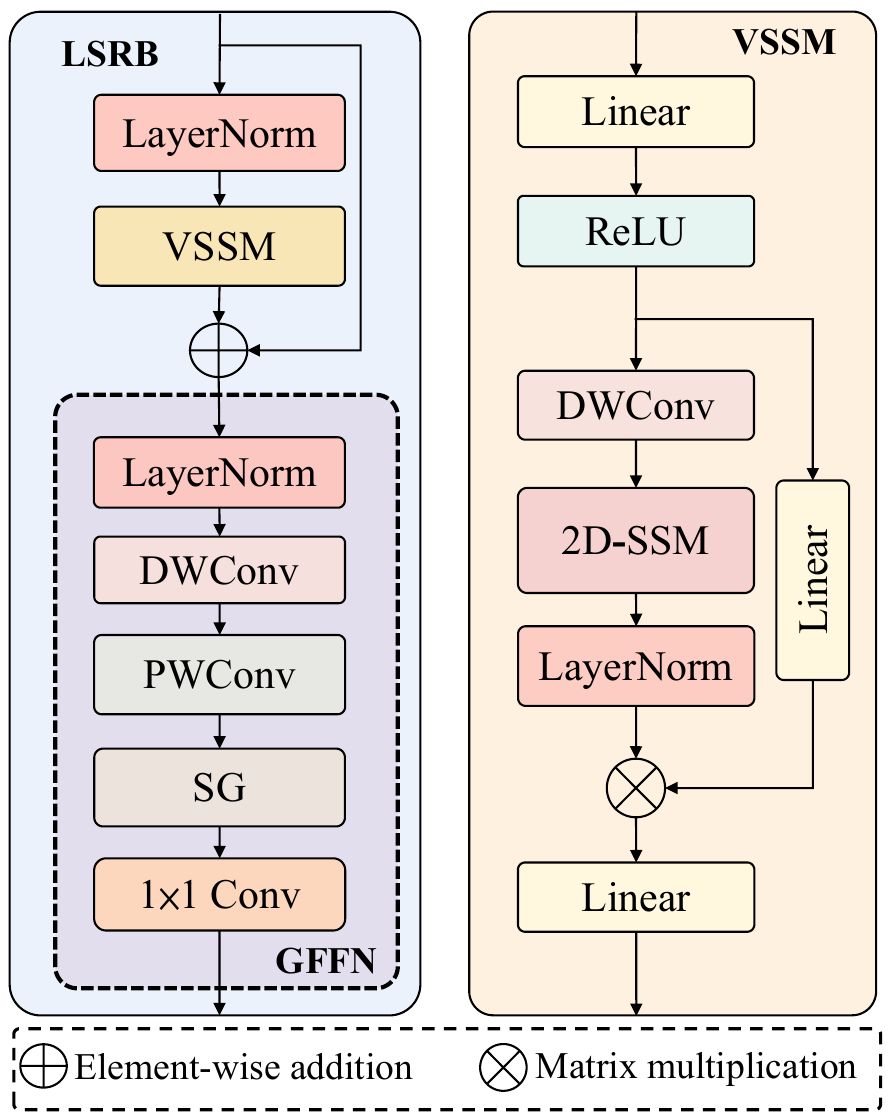}
	\caption{Architecture of the Low-frequency Structure Restoration Block (LSRB) and the Vision State Space Module (VSSM). 
 	}
 	\label{fig:fig11}
 \end{figure}

\subsubsection{Low-frequency Structure Restoration Block} According to frequency domain analysis, low-frequency components primarily encode the global structural attributes of the image, including background and color cues. Owing to Mamba’s strong global modeling capabilities and its efficient linear computational complexity, it is particularly well-suited for handling such components. Therefore, we employ Mamba as the principal module for reconstructing low-frequency image features. Specifically, we introduce a Low-frequency Structure Restoration Block (LSRB) to capture and model global information flow from low-frequency features, thereby enhancing spatial-domain representation, as shown in Fig. \ref{fig:fig11} . Given an input low-frequency feature map $F_{L}' \in \mathbb{R}^{H/2\times W/2 \times C}$, we first apply Layer Normalization, followed by a Visual State Space Module to effectively capture long-range dependencies within global contextual information. In addition, to further improve the efficiency of channel-wise information propagation, we integrate a Gated Feed-Forward Network. The overall computation process can be formulated as:
\begin{align}
    Z &= \mathit{VSSM}\big(\mathit{LN}(F_{L}')\big) + \beta F_L ,\\
    F_{L}'' &= \mathit{GFFN}(Z) + \gamma Z,
\end{align}
where $VSSM(\cdot)$ and $GFFN(\cdot)$ denote the Visual State Space Module and Gated Feed-Forward Network, respectively. $LN(\cdot)$ represents the layer normalization operation. $Z$ is the intermediate representation output by the VSSM, which is subsequently refined by the GFFN. The learnable parameters $\beta$ and $\gamma$ serve as scaling factors to adaptively modulate the feature distribution.

\begin{figure}[t] 
    \centering
    \includegraphics[width=1.0\linewidth]{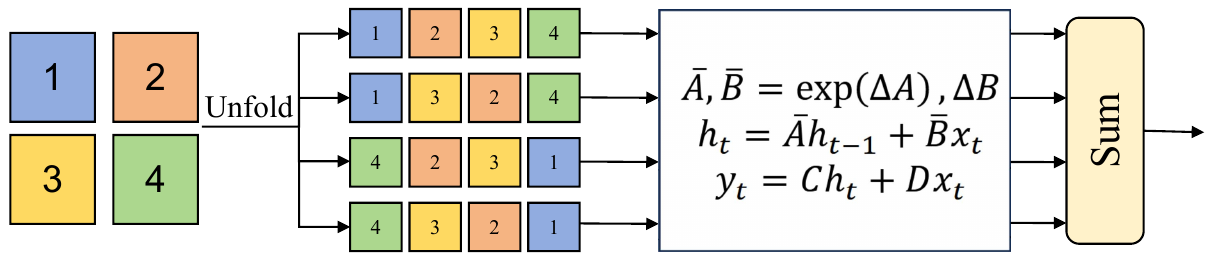}
	\caption{Schematic diagram of the four-directional scanning mechanism in 2D-SSM for spatial state modeling. Pixels are sequentially scanned along four axial pathways and processed through spatial state transformations to effectively capture comprehensive long-range dependencies.
 	}
 	\label{fig:fig4}
 \end{figure}

\textbf{Vision State Space Module.} Building on the efficiency of Mamba in modeling long-range dependencies with linear computational complexity, we introduce the Vision State Space Module (VSSM) to enhance the modeling of global low-frequency feature. As illustrated in Fig. \ref{fig:fig11} right, VSSM leverages state-space formulations to capture extended spatial dependencies efficiently. The overall calculation process is expressed by the following formula:
\begin{align}
    X_1 & = \mathit{SiLU}\big(\mathit{DWConv}(\mathit{Linear}(X))\big) \label{eq:path1}, \\
    X_2 &= \mathit{LN}\big(\mathit{2D\text{-}SSM}(X_1)\big) \label{eq:path2}, \\
    X_{\mathit{out}} &= \mathit{Linear}\Big(X_2 \odot \big(\mathit{SiLU}(\mathit{Linear}(X))\big)\Big),
    \label{eq:fusion}
\end{align}
where DWConv represents depthwise separable convolution, while 2D-SSM refers to the two-dimensional selective scan module. The 2D-SSM transforms 2D features into sequential data via four directional scanning paths (Fig. \ref{fig:fig4}), performs state-space modeling to capture long-range dependencies, and reconstructs the output into spatial features for global context modeling. The symbol $\odot$ denotes element-wise (Hadamard) multiplication.

% The processing pipeline begins with a depthwise convolution followed by a SiLU activation function (Eq. \ref{eq:path1}) for local pattern extraction. The resulting features are then fed through a 2D-Selective Scanning Module (2D-SSM) with LayerNorm (Eq. \ref{eq:path2}) to model global dependencies, producing intermediate denoted as features $X_1$. In the final stage, feature fusion is performed via a Hadamard product between the global features $X_2$ and the activated output from the initial expansion layer (Eq.~\ref{eq:fusion}). This is followed by a linear projection layer that restores the original spatial dimensions in the final output $X_{\mathrm{out}}$. Here, $DWConv(\cdot)$ and $Linear(\cdot)$ represent depthwise convolution and linear projection operations, respectively, while $\odot$ denotes the element-wise Hadamard product.

\textbf{Gate Feed-Forward Network.} In our design, the Gate Feed-Forward Network (GFFN) utilizes a nonlinear gating mechanism to regulate information flow, enabling each channel to be more effectively characterized. The operation of the GFFN is defined as:
\begin{equation}
    \text{\small $Z_{\mathit{out}} = \mathit{DWConv}\Big(\delta_{\mathit{SG}}\big(\mathit{DWConv}(\mathit{PWConv}(\mathit{LN}(Z_{\mathit{in}}))\big)\Big)$},
\end{equation}
where $\delta_{SG}(\cdot)$ denotes the simple gate mechanism function \cite{ren2016gated}. This mechanism splits the input tensor along the channel dimension into two feature maps, $\textbf{F}_1, \textbf{F}_2 \in \mathcal{R}^{H\times W\times \frac{C}{2}}$. 
% The output is then computed using a nonlinear gating operation:
% \begin{equation}
% \delta_{SG}(\textbf{F}) = GELU(\textbf{F}_1)\odot \textbf{F}_2, 
% \end{equation}
% where $GELU(\cdot)$ denotes the Gaussian Error Linear Unit activation function. $Z_{\text{in}}$ and $Z_{\text{out}}$ refer to the input and output of the GFFN, respectively.
% \textbf{2D Selective Scan Module.} To address the inherent limitations of Mamba in processing non-sequential image data, which is originally designed for causal processing in sequential NLP tasks, we introduce the 2D Selective Scan Module (2D-SSM) for effective spatial information modeling.
% Following the methodology in \cite{guo2025mambair}, 2D-SSM systematically transforms 2D feature maps into linear sequences through four directional scanning paths: (1) top-left to bottom-right, (2) bottom-right to top-left, (3) top-right to bottom-left, and (4) bottom-left to top-right (see Fig. \ref{fig:fig4}). Each directional sequence undergoes discrete state-space modeling to capture long-range contextual dependencies. Subsequently, the processed sequences are aggregated and transformed back into the original 2D spatial configuration, enabling effective global context modeling while maintaining structural integrity.
\subsubsection{High-frequency Detail Enhancement Block} 
\begin{figure}[t] 
    \centering
    \includegraphics[width=1.0\linewidth]{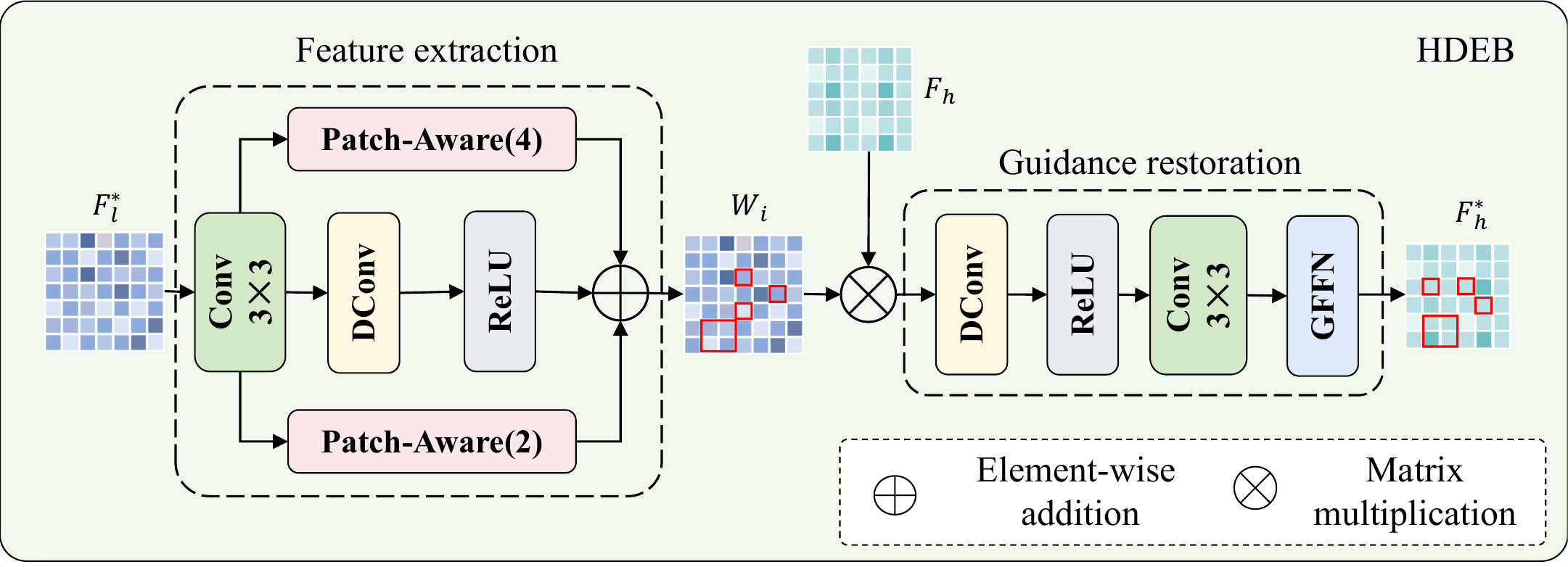}
	\caption{
        Architecture of the High-frequency Detail Enhancement Block (HDEB). HDEB first extracts patch-level features from the low-frequency input to localize fine details. It then performs targeted enhancement, enabling precise recovery of high-frequency components and intricate texture details in the dehazed image.
 	}
 	\label{fig:fig5}
 \end{figure}

High-frequency information, as revealed by Laplace frequency analysis, typically encodes fine-grained structural elements such as edges and textures. Convolutional neural networks (CNNs), with their inherent strength in localized pattern extraction, are particularly well-suited for modeling such details. To this end, we propose a High-Frequency Detail Enhancement Block (HDEB) designed to effectively restore local details in hazy images, as illustrated in Fig. \ref{fig:fig5}. To further improve high-frequency reconstruction, we integrate low-frequency features to guide the enhancement process. Specifically, HDEB leverages Pixel-wise Attention (PA) \cite{xu2024hcf} to extract and refine features from localized patch blocks, thereby emphasizing structurally salient regions. Each HDEB comprises two primary stages: local feature extraction and guided restoration. In the extraction phase, patch blocks of sizes 2$\times$2 and 4$\times$4 are sampled from the low-frequency feature map. These patches are then processed to derive spatially localized cues. During the fusion stage, an attention mechanism is applied to amplify significant features, enabling the network to effectively restore fine textures and intricate structures in the high-frequency domain. The specific process can be expressed as:
\begin{align}
    &\text{\small $W_i = \mathcal{A}(\mathit{ReLU}\big(\mathit{DConv}(F_l^*)\big) 
           + \mathit{PA}^{4}(F_l^*)
           + \mathit{PA}^{2}(F_l^*)$} \label{eq:low}, \\
    &\text{\small $F_{h}^{\ast} = \mathit{GFFN}\Big(\mathit{Conv}\big(\mathit{ReLU}(
           \mathit{DConv}(W_i \odot F_{h}))\big)\Big)$} \label{eq:high},
\end{align}
where $F_{l}^*$, $F_{h}$, and $F_{h}^*$ represent the input low-frequency features, the input high-frequency features, and the enhanced high-frequency output, respectively. $PA^{2}()$ and $PA^{4}()$ denote the pixel-wise attention operations applied to extract local features from 2$\times$2 and 4$\times$4 patch blocks.  The attention weights $W_i$ are computed to modulate the fusion of local features with the corresponding high-frequency components, thereby reinforcing salient structural details. To further enhance naturalness and fidelity in the restored image, the final local detail refinement is performed using a standard DConv-ReLU-Conv structure.

%\subsection{Optimization Objectives}
\subsection{Loss Functions}
Most image dehazing methods rely on pixel-wise loss functions to optimize model performance. Although $L_2$ loss is commonly employed, prior studies \cite{zhao2016loss, lim2017enhanced} have demonstrated that the $L_1$ loss tends to yield better perceptual quality and improved metrics such as PSNR and SSIM for image restoration tasks. Based on this observation, we adopt the standard $L_1$ loss to supervise the training of our network. 

To further encourage the network to retain both global structures and fine-grained details, especially in the frequency domain, we incorporate an additional frequency-aware loss inspired by \cite{jiang2021focal}. Specifically, we perform a discrete Laplace-frequency transform (LFT) on both the predicted and ground truth images and compute the $L_1$ loss in the frequency domain. The total loss function is formulated as a weighted sum of spatial and frequency losses:
\begin{equation}
      \mathcal{L}_{total} = \mathcal{L}_{recon} + \lambda \cdot \mathcal{L}_{freq},
\end{equation}
where $\lambda$ is a weighting coefficient that balances the two components. Following the experimental setup of Wave-Mamba~\cite{wang2024wave}, we set the value of $ \lambda$ to 0.1.
The spatial reconstruction loss $\mathcal{L}_{recon}$ is defined as:
\begin{equation}
\mathcal{L}_{recon} = \frac{1}{N} \sum_{i=1}^{N} \left\| I_i^{\text{gt}} - \mathcal{F}(I_i^{\text{haze}}) \right\|_1.
\end{equation}

The frequency-domain reconstruction loss $\mathcal{L}_{freq}$ is expressed as:
\begin{equation}
\mathcal{L}_{freq} = \frac{1}{N} \sum_{i=1}^{N} \left\| \mathcal{F}_{\text{LFT}}(I_i^{\text{gt}}) - \mathcal{F}_{\text{LFT}}(\mathcal{F}(I_i^{\text{haze}})) \right\|_1,
\end{equation}
where $\mathcal{F}$ denotes the dehazing network, $I^{\text{gt}}$ and $I^{\text{haze}}$ represent the ground truth and hazy input images, respectively, and $\mathcal{F}_{\text{LFT}}$ indicates the discrete Laplace-frequency transform.

Empirically, we find that this composite loss formulation effectively enhances the capacity of our method to reconstruct high-quality dehazed images, preserving both spatial fidelity and frequency-domain consistency.

% \begin{equation}
%       \mathcal{L}_{\text{total}} = \mathcal{L}_{\text{L1}} + \lambda \cdot \mathcal{L}_{\text{freq}}
% \end{equation}
% \begin{equation}
% \mathcal{L}_{\text{L1}} = \frac{1}{N} \sum_{i=1}^{N} \left\| I_i^{\text{gt}} - \mathcal{F}(I_i^{\text{haze}}) \right\|_1
% \end{equation}
% \begin{equation}
% \mathcal{L}_{\text{freq}} = \frac{1}{N} \sum_{i=1}^{N} \left\| \mathcal{F}_{\text{DFT}}(I_i^{\text{gt}}) - \mathcal{F}_{\text{DFT}}(\mathcal{F}(I_i^{\text{haze}})) \right\|_1
% \end{equation}

% Here, \( \mathcal{F} \) represents the dehazing network, \( I^{\text{gt}} \) is the ground truth image, and \( I^{\text{haze}} \) is the hazy input. \( \lambda \) is a balancing weight between spatial and frequency losses. We empirically find that this combined loss encourages better detail recovery and frequency consistency in dehazed results.

\section{Experiments}
%In this section, we conduct comprehensive experiments to evaluate the performance of Laplace-Mamba in comparison with its competitors. We begin by introducing the benchmark dataset and detailing the implementation setup used in our experiments. Next, we provide an in-depth description of the state-of-the-art algorithms and image quality assessment metrics employed for comparison. Subsequently, we present both quantitative and qualitative experimental results to highlight the superiority of our proposed network. Finally, we conduct an ablation study to analyze the contribution and effectiveness of each component in Laplace-Mamba.
In this section, we present a comprehensive evaluation of the proposed Laplace-Mamba framework. We first describe the benchmark datasets and implementation details used in our experiments. Next, we outline the state-of-the-art algorithms selected for comparison and the image quality assessment metrics employed. We then provide both quantitative and qualitative results, demonstrating the superior performance of Laplace-Mamba over existing methods. Finally, we perform an ablation study to systematically assess the impact of each component in our framework.

\begin{table}[t]
	\centering
	\footnotesize
	\caption{Quantitative comparisons (Average PSNR/SSIM) of SOTA dehazing methods on the Haze4K \cite{song2021haze4k} validation set. $\uparrow$ indicates that higher values are better. Boldface values highlight the best results}
	\label{Tab:1}
	\begin{threeparttable}
% 		\normalsize
		\centering
		\setlength{\tabcolsep}{2mm}{
			\begin{tabular}{lccc}
				\toprule
				\multirow{1}{*}{Method}&
				\multirow{1}{*}{Publication}&
				\multirow{1}{*}{PSNR$\uparrow$}&
				\multirow{1}{*}{SSIM$\uparrow$}\cr
				\midrule
				AOD-Net \cite{li2017aod} & ICCV'17 & 17.15 & 0.830  \cr
				GridDehazeNet \cite{liu2019griddehaze} & ICCV'19 & 23.29 & 0.848  \cr
				MSBDN \cite{dong2020msbdn} & ICCV'20 & 22.99  &  0.854   \cr
				FFA-Net \cite{qin2020feature} & AAAI‘20 & 26.96 & 0.966  \cr
				DMT-Net \cite{ullah2021dmt} & MM'21 &  28.53 & 0.960  \cr
                Dehazeformer \cite{cai2016dehazenet} & TIP'23 & 30.75 &0.981   \cr
                LMHaze  \cite{zhang2022lmhaze} &MM'24  & 30.51  & 0.972  \cr
			  DEA-Net \cite{bai2024dea} & TIP'24 & 34.25 & 0.987  \cr
				ConvIR-L \cite{cui2024revitalizing} & TPAMI'24 & 34.50 & 0.988  \cr
                \midrule
				Laplace-Mamba & Ours & \textbf{35.70} &
				\textbf{0.991} \cr
				\bottomrule
			\end{tabular}
		}
	\end{threeparttable}
\end{table}

\subsection{Implementation Details}
\textbf{Datasets.} 
%Our experiments employ synthetic data sets to evaluate the performance of the Laplace-Mamba model, mainly using the Haze4K dataset \cite{song2021haze4k}, which contains 5,040 pairs of hazy/haze-free 4K-resolution ($3840\times2160$) images with varying haze densities, including 3,925 training pairs (3,545 outdoor scenes and 1,495 indoor scenes) and 1,115 testing pairs with strict scene-level separation to prevent data leakage. For comparative analysis, We further evaluate on the LMHaze dataset \cite{zhang2022lmhaze}, a synthetic benchmark containing 4,000 hazy/haze-free pairs (3,000 training and 1,000 testing) specifically designed for low-light haze conditions with randomized atmospheric parameters ($\beta\in[0.6,2.5]$, $A\in[0.7,1]$) and simulated illumination ($\mathcal{L}\in[0.1,0.3]$). The dataset features 12 scene categories with per-pixel transmission maps, 14-bit RAW data, and a unique 500-image real-world validation subset, employing strict non-overlapping scene splits to prevent evaluation bias.
We evaluate the proposed Laplace-Mamba model on three benchmark datasets: two widely recognized synthetic collections (Haze4K \cite{song2021haze4k} and LMHaze \cite{zhang2022lmhaze}) and one real-world dataset (O-Haze \cite{ancuti2018haze}). Haze4K contains 5,040 high-resolution (4K, 3840$\times$2160) image pairs, including 3,925 training pairs (3,545 outdoor and 1,495 indoor scenes) and 1,115 testing pairs, ensuring strict scene-level separation to prevent data leakage. LMHaze focuses on low-light hazy conditions, offering 4,000 image pairs (3,000 for training and 1,000 for testing) with randomized atmospheric parameters ($\beta \in [0.6, 2.5]$, $A \in [0.7, 1]$) and simulated illumination ($\mathcal{L} \in [0.1, 0.3]$). To bridge the synthetic-to-real gap, we additionally validate on O-Haze, a benchmark dataset containing 45 professionally captured outdoor scene pairs generated through physical haze simulation. Unlike synthetic datasets, O-Haze captures authentic non-uniform haze distributions using haze machines, with each scene containing synchronized multi-intensity haze variations and accompanying calibrated atmospheric measurements for reference. All three datasets maintain strict non-overlapping scene splits, with LMHaze and O-Haze further providing auxiliary data: LMHaze includes 12 scene categories, per-pixel transmission maps, and 14-bit RAW data plus a 500-image real-world validation subset, while O-Haze offers coarse depth estimates for physics-based analysis.

\textbf{Architecture.} 
%Our network architecture employs a hierarchical design with layer-specific configurations. The encoder and decoder components utilize [1, 1, 2, 4] Local Spatial Residual Blocks (LSRBs) and [1, 1, 1, 2] Hierarchical Dense Building Blocks (HDBBs) at each corresponding layer. For the loss function, we implement a weighted combination of $L_1$ loss (weight=0.9) and frequency domain loss (weight=0.1) to optimize both spatial and spectral characteristics. To enhance feature representation, we set the channel dimension C to 32 throughout the network.
The proposed Laplace-Mamba framework adopts a U-Net-like architecture comprising seven network layers with three downsampling and three upsampling stages. The first three layers incorporate three core components: (1) Laplace-frequency transformation module, (2) multi-domain fusion module, and (3) Laplace-Mamba module, while the subsequent four layers only include the Laplace-frequency transformation module and Laplace-Mamba module. Furthermore, each Laplace-Mamba module integrates $M$ Low-frequency Structure Restoration Blocks (LSRBs) and $N$ High-frequency Detail Enhancement Blocks (HDEBs), with layer-specific configurations $M = [1, 1, 2, 4, 2, 1, 1]$ and $N = [1, 1, 1, 2, 1, 1, 1]$ across the seven-layer architecture.

\textbf{Training Details.}
%All experiments were implemented using PyTorch 1.7 on a system with an Intel i9-12900KF CPU and an NVIDIA RTX 3090 GPU. The network was optimized using the Adam optimizer \cite{kingma2014adam} with a batch size of 8. The number of LSRB for each layer of the network encoder and decoder was [1, 2, 4], and the number of HDEB was [1, 1, 1]. Note that the number of headers is 8 and the number of channels $C$ is 32. The initial learning rate is $5\times10^{-4} $, which is gradually reduced to $1\times10^{-7}$ by cosine annealing for a total of 500k iterations. We use random rotations of 90, 180, and 270, random flips, and random cropping to obtain enhanced training data of $256 \times 256$ size. To limit the training of Laplace-Mamba, we used the L1 loss function.
All experiments are implemented using PyTorch 1.7 on a system equipped with an Intel i9-12900KF CPU and an NVIDIA RTX 3090 GPU. The network is optimized using the Adam optimizer \cite{kingma2014adam} with a batch size of 8. The learning rate is initialized at $5\times10^{-4}$ and gradually decays to $1\times10^{-7}$ following a cosine annealing schedule over 500k iterations. Data augmentation includes random rotations (90°, 180°, 270°), horizontal and vertical flips, and random cropping to 256$\times$256 patches.

\textbf{Evaluation Metrics.} 
%The well-known full-referenced image quality assessment metrics Peak Signal to Noise Ratio (PSNR) \cite{dong2015image} and Structural Similarity Index (SSIM) \cite{wang2004image} are adopted to evaluate the performance of Laplace-Mamba and other dehazing algorithms quantitatively. Specifically, PSNR is used to calculate the distinction between the restored images and the ground-truth images in the corresponding pixels, while SSIM is employed to evaluate image similarity in terms of brightness, contrast, and structure. Larger values of these two metrics commonly indicate better results. In addition, we compute the computational complexity of the models using FLOPs\cite{molchanov2019taylor}, which quantifies the computational complexity of deep learning models by counting the number of floating-point arithmetic operations (\emph{e.g.}, additions, multiplications) required for a single forward pass. It is a standardized metric for comparing the efficiency of models from different architectures.
We employ two widely recognized image quality assessment metrics for comprehensive evaluation of Laplace-Mamba and its competitors: Peak Signal-to-Noise Ratio (PSNR) \cite{dong2015image}, Structural Similarity Index (SSIM) \cite{wang2004image}. Specifically, PSNR measures the pixel-wise fidelity between restored and ground-truth images, where higher values indicate superior restoration quality. SSIM quantifies image similarity by evaluating brightness, contrast, and structural integrity, with higher values representing better preservation of image quality. 
% Furthermore, we assess computational efficiency using Floating Point Operations Per Second (FLOPs) \cite{molchanov2019taylor}, which accurately measures model complexity by counting the number of arithmetic operations (\emph{e.g.}, additions, multiplications) during a single forward pass. This standardized metric enables fair comparisons of computational cost across different architectures.

\begin{table*}[t]
	\centering
    \large
	\caption{Quantitative comparisons (PSNR/SSIM) between Laplace-Mamba and 8 SOTA dehazing approaches on the LMHaze dataset}
	\label{Tab:2}
	\begin{threeparttable}
        \footnotesize
        \centering
        \setlength{\tabcolsep}{14pt} % Increased column spacing
        \renewcommand{\arraystretch}{1.0} % Increased row spacing
			\begin{tabular}{lcccccccc}
				\toprule
				\multirow{2}{*}{Method}&
				\multirow{2}{*}{Publication}&
				\multicolumn{2}{c}{LMHaze}&
				\multicolumn{2}{c}{LMHaze indoor}&
                \multicolumn{2}{c}{LMHaze outdoor}\cr
				 \cmidrule(lr){3-4} \cmidrule(lr){5-6} \cmidrule(lr){7-8}&
				 & PSNR$\uparrow$ & SSIM$\uparrow$   &PSNR$\uparrow$ & SSIM$\uparrow$  & PSNR$\uparrow$ & SSIM$\uparrow$  \cr
				\midrule
            DehazeNet & TIP'16 & 12.71 & 0.609 & 12.97 & 0.629 & 12.45 & 0.589 \\
            AOD-Net & ICCV'17 & 14.83 & 0.607 & 15.07 & 0.659 & 14.59 & 0.554 \\
            GridDehazeNet & ICCV'19 & 15.93 & 0.673 & 15.70 & 0.664 & 16.16 & 0.681 \\
            FFA-Net & AAAI'20 & 16.30 & 0.696 & 16.46 & 0.676 & 16.13 & 0.715 \\
            Dehamer & CVPR'22 & 15.76 & 0.574 & 16.00 & 0.578 & 15.51 & 0.569 \\
            DehazeFormer & TIP'23 & 17.86 & 0.763 & 18.04 & 0.757 & 17.67 & 0.769 \\
            MambaIR & ECCV'24 & 17.95 & 0.753 & 17.93 & 0.750 & 17.96 & 0.755 \\
            LMHaze & MM'24 & 18.52 & 0.782 & 18.56 & 0.790 & 18.48 & 0.774  \\
            \midrule
            Laplace-Mamba & Ours & \textbf{20.60} & \textbf{0.814} & \textbf{20.51} & \textbf{0.803} & \textbf{20.66} & \textbf{0.821} \\
				\bottomrule
			\end{tabular}
        %}
	\end{threeparttable}
\end{table*}

\begin{table}[t]
	\centering
	\footnotesize
	\caption{Quantitative comparisons (Average PSNR/SSIM) of SOTA dehazing methods on the O-Haze \cite{ancuti2018haze} validation set}
	\label{Tab:6}
	\begin{threeparttable}
		\centering
		\setlength{\tabcolsep}{2mm}{
			\begin{tabular}{lccc}
				\toprule
				\multirow{1}{*}{Method}&
				\multirow{1}{*}{Publication}&
				\multirow{1}{*}{PSNR$\uparrow$}&
				\multirow{1}{*}{SSIM$\uparrow$}\cr
				\midrule
                AOD-Net \cite{li2017aod} & ICCV'17 & 18.19 & 0.6823  \cr
				GridDehazeNet \cite{liu2019griddehaze} & ICCV'19 & 20.05 & 0.7362  \cr
				FFA-Net \cite{qin2020feature} & AAAI‘20 & 23.34 & 0.8084  \cr
                Dehamer \cite{guo2022dehamer} & CVPR‘22  &24.36  & 0.8089  \cr
                Dehazeformer-L \cite{cai2016dehazenet} & TIP'23 & 25.25 &0.8206   \cr
                DEA-Net \cite{bai2024dea} & TIP'24 & 25.54 & 0.8196  \cr
			  OKNet \cite{bai2024dea} & TCSVT'24 & 25.62 & 0.8528  \cr
				ConvIR-L \cite{cui2024revitalizing} & TPAMI'24 & 26.09 & 0.8552  \cr
                \midrule
				Laplace-Mamba & Ours & \textbf{26.51} &
				\textbf{0.8582}\cr
				\bottomrule
			\end{tabular}
		}
	\end{threeparttable}
\end{table}

\subsection{Comparison with State-of-the-art Methods} 
%In this section, we evaluate the proposed Laplace-Mamba against eight representative SOTA dehazing approaches in both quantitative and qualitative assessments, including DehazeNet \cite{cai2016dehazenet}, AOD-Net \cite{li2017aod}, GridDehaze \cite{liu2019griddehaze}, MSBDN \cite{dong2020msbdn}, FFA-Net \cite{qin2020feature}, DMT-Net \cite{ullah2021dmt}, DEA-Net \cite{bai2024dea}, ConIR\cite{cui2024revitalizing}, and a more recent Mamba-based dehazing framework LMHaze \cite{zhang2022lmhaze}. We present both qualitative and quantitative comparisons of our Laplace-Mamba model against current state-of-the-art methods. To fully demonstrate the strengths of our approach, we perform complementary experiments on the dense haze dataset, comparing LMHaze Indoor with LMHaze Outdoor. Our evaluation employs three metrics: PSNR, SSIM, and LPIPS \cite{zhang2018unreasonable}. PSNR quantifies the fidelity of reconstructed images by comparing pixel intensity differences, SSIM assesses image quality by analyzing brightness, contrast, and structural similarity, and LPIPS measures perceptual similarity using deep learning-based models.

We compare Laplace-Mamba against eight state-of-the-art (SOTA) dehazing methods across quantitative and qualitative assessments: DehazeNet \cite{cai2016dehazenet}, AOD-Net \cite{li2017aod}, GridDehaze \cite{liu2019griddehaze}, MSBDN \cite{dong2020msbdn}, FFA-Net \cite{qin2020feature}, DMT-Net \cite{ullah2021dmt}, DEA-Net \cite{bai2024dea}, ConIR \cite{cui2024revitalizing}, and the recent Mamba-based dehazing framework LMHaze \cite{zhang2022lmhaze}. Our evaluation spans both quantitative and qualitative perspectives. For quantitative evaluation, we compute PSNR and SSIM metrics across three primary benchmarks, offering a comprehensive assessment of image fidelity and perceptual quality. On the qualitative front, we provide visual comparisons to underscore the efficacy of Laplace-Mamba in recovering intricate details and preserving natural textures. To thoroughly evaluate the generalization capability of our approach, we conduct extensive experiments on both synthetic (Haze4K and LMHaze) and real-world (O-Haze) hazy conditions. This systematic analysis demonstrates the superior restoration accuracy, robustness, and adaptability of our method across diverse degradation scenarios, highlighting its practical applicability.

\textbf {Quantitative Evaluation.}
% As exhibited in TABLE \ref{Tab:1}, we report the average PSNR, SSIM, and Gfloat values of nine dehazing algorithms on the Haze4K validation set for quantitative evaluation. For a fair comparison, all compared methods are retrained on the Haze4K training set, following the settings in their papers. Clearly, our Laplace-Mamba achieves the best performance in PSNR, SSIM, and FLOPs, and outperforms the SOTA dehazing approaches by a huge margin. The quantitative results fully verify the effectiveness of Laplace-Mamba on the image dehazing tasks.
We quantitatively evaluate the performance of Laplace-Mamba in comparison with several state-of-the-art (SOTA) dehazing algorithms on three benchmark datasets (\textit{i.e.}, Haze4K \cite{song2021haze4k}, LMHaze \cite{zhang2022lmhaze}, and O-Haze \cite{ancuti2018haze} datasets). 
Table \ref{Tab:1} reports the average PSNR and SSIM values on the Haze4K validation set for nine representative dehazing methods alongside our proposed Laplace-Mamba. As illustrated, Laplace-Mamba achieves the best performance, highlighting its superior restoration quality under synthetic degradation.
To further validate the effectiveness of our approach, we conduct extensive experiments on both synthetic (LMHaze) and real-world (O-Haze) datasets. As demonstrated in Table \ref{Tab:2}, comprehensive comparisons on the LMHaze dataset, which includes both indoor and outdoor hazy scenes, demonstrate that our method consistently outperforms all competing approaches across all metrics. Specifically, Laplace-Mamba achieves the highest PSNR and SSIM scores, affirming its exceptional capability in pixel-level fidelity restoration and perceptual quality preservation.
Additionally, Table~\ref{Tab:6} presents results on the O-Haze dataset, comprising real-world hazy images. The test images are uniformly cropped to a resolution of 1600$\times$1200 pixels for standardized processing. As observed, our model demonstrates substantial improvements over existing methods, showcasing its robustness in practical scenarios with uncontrolled environmental factors. The consistent superiority of Laplace-Mamba across Haze4K, LMHaze, and O-Haze conclusively validates its generalization capacity and effectiveness in diverse haze conditions. These results underscore the strong potential of our approach to deliver high-fidelity dehazing under both synthetic and real-world scenarios.

\begin{figure*}[t] 
        \centering
 	\includegraphics[width=0.95\linewidth]{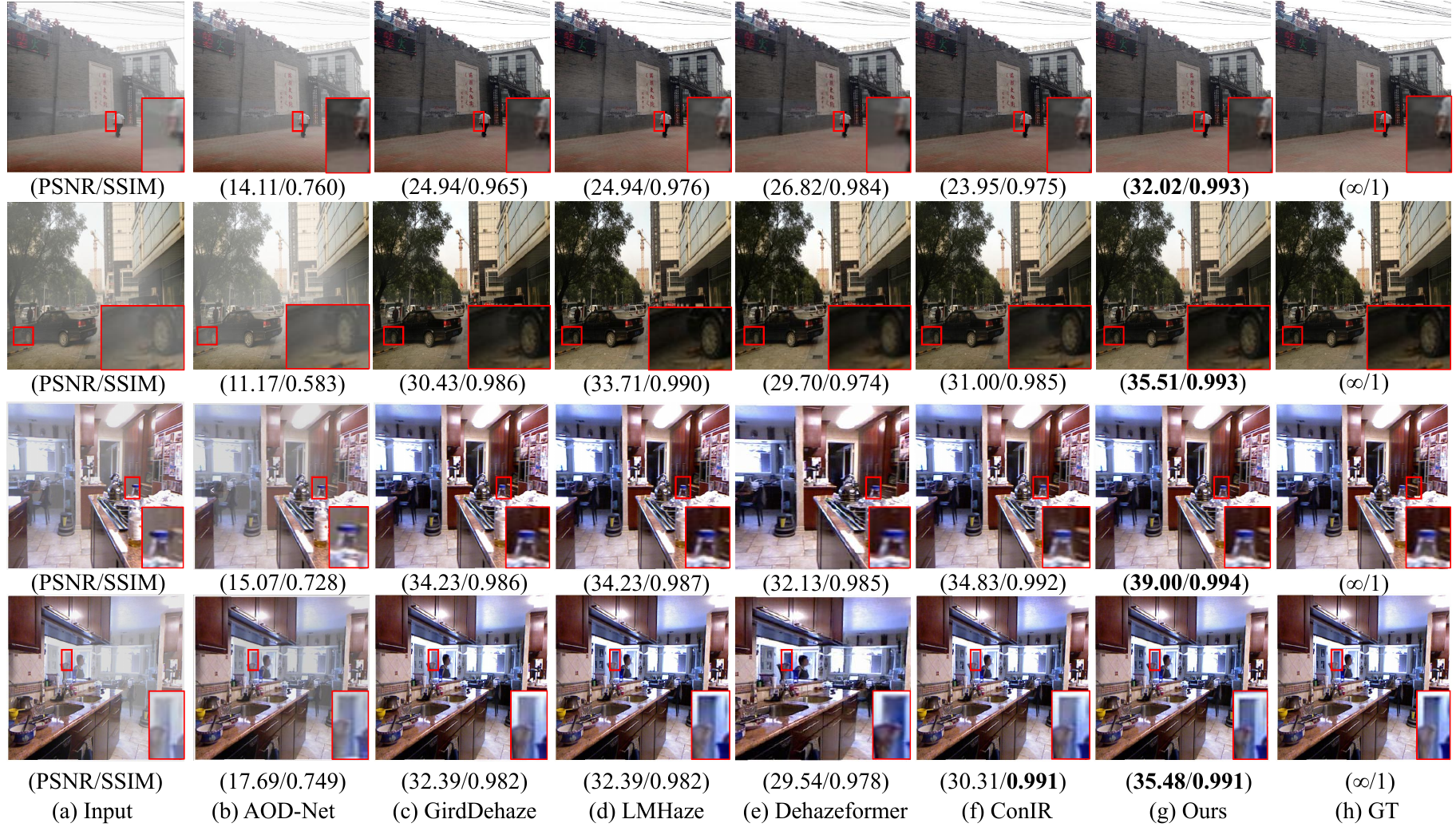}
 	\caption{ Qualitative comparisons on the Haze4K validation set. From (a) to (g): (a) input hazy images, and the dehazing results of (b) AOD-Net \cite{li2017aod}, (c) GridDehazeNet \cite{liu2019griddehaze}, (d) LMHaze \cite{zhang2022lmhaze}, (e) DehazeFormer \cite{song2023dehazeformer}, (f) our proposed Laplace-Mamba, respectively, and (g) the ground-truth image. Our Laplace-Mamba produces natural and clearer haze-free images.}
 	\label{fig:fig6}
\end{figure*}

\begin{figure*}[!h] 
  \centering
	\includegraphics[width=0.95\linewidth]{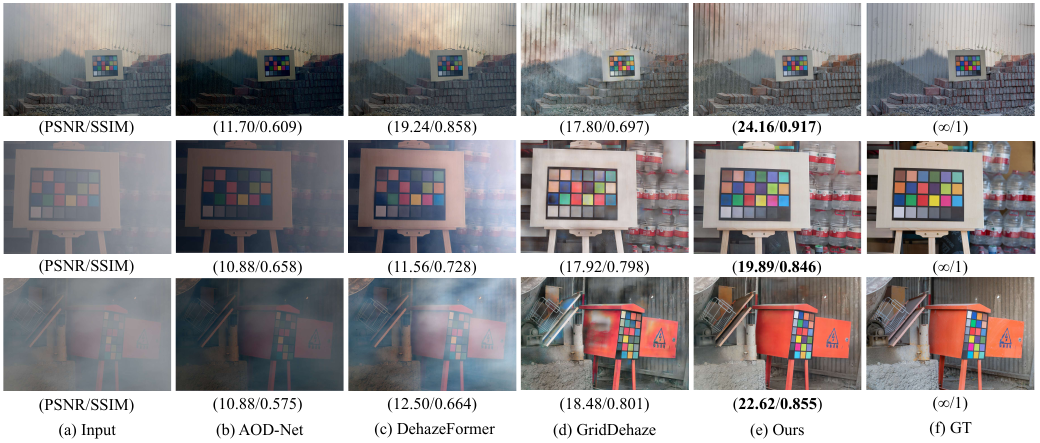}
	\caption{Qualitative comparison on the LMHaze\cite{zhang2022lmhaze} dataset. From (a) to (e): (a) input hazy images, (b) the dehazing result of AOD-Net \cite{li2017aod}, (c) DehazeFormer \cite{song2023dehazeformer}, (d) GridDehazeNet \cite{liu2019griddehaze}, and (e) our proposed Laplace-Mamba. Laplace-Mamba demonstrates superior haze removal, effectively preserving fine details while maintaining natural color balance.}
	\label{fig:fig7}
\end{figure*}

\textbf{Qualitative Evaluation.} 
% As illustrated in Fig. \ref{fig:fig6}, we conducted a systematic comparative analysis using six representative samples from the Haze4K\cite{song2021haze4k} dataset to demonstrate the performance variations among different dehazing algorithms. The AOD\cite{li2017aod} algorithm exhibits noticeable haze residue in processed images, indicating significant limitations in complete haze removal. Both GridDehaze\cite{liu2019griddehaze} and LMHaze\cite{zhang2022lmhaze} introduce apparent artifacts, manifesting as localized color fading and overall luminance reduction. While the Dehazeformer\cite{song2023dehazeformer} achieves certain dehazing effects, magnified examination reveals severe image blurring that substantially compromises visual quality. The ConIR\cite{cui2024revitalizing} algorithm demonstrates relatively superior performance overall, yet still suffers from residual haze patches in specific regions, constraining its effectiveness. In contrast, our proposed Laplace-Mamba algorithm demonstrates remarkable advantages: Firstly, it effectively eliminates various degrees of haze interference, including dense haze conditions. Secondly, it maintains excellent image quality throughout the dehazing process. More notably, compared with other state-of-the-art algorithms, Laplace-Mamba produces dehazed images with more natural color restoration and more complete structural detail preservation, thereby achieving visually more authentic dehazing results. These advantages endow the Laplace-Mamba algorithm with outstanding performance in practical applications.
As shown in Fig. \ref{fig:fig6}, we conduct a comprehensive qualitative analysis using four representative samples from the Haze4K \cite{song2021haze4k} dataset to evaluate the performance differences among various dehazing methods. The AOD-Net \cite{li2017aod} method leaves noticeable haze residues in the processed images, highlighting its limitations in complete haze removal. Both GridDehaze \cite{liu2019griddehaze} and LMHaze \cite{zhang2022lmhaze} introduce distinct artifacts, including localized color fading and reduced overall luminance. While DehazeFormer \cite{song2023dehazeformer} achieves partial haze removal, closer inspection reveals significant image blurring, adversely affecting visual quality. The ConvIR \cite{cui2024revitalizing} method demonstrates relatively superior performance compared to the aforementioned approaches, but residual haze patches remain in specific areas, limiting its effectiveness. In contrast, our proposed Laplace-Mamba method exhibits several significant advantages. First, it effectively removes varying degrees of haze, even under dense haze conditions. Second, it consistently maintains high image quality throughout the dehazing process. Notably, compared with other state-of-the-art methods, Laplace-Mamba produces dehazed images with more natural color restoration and superior structural detail preservation, resulting in visually more realistic dehazing outcomes. These qualities make the Laplace-Mamba method particularly suitable for practical applications.

%To further evaluate the haze removal capability of Laplace-Mamba, we present the restoration results for a challenging hazy image in Fig. \ref{fig:fig6}. As demonstrated in the figure, AOD-Net fails to eliminate the haze and exhibits limited detail recovery performance; GridDehazeNet recovers substantially more details than AOD-Net, yet produces an overly bright and unnatural-looking result; DehazeFormer shows superior detail preservation compared to the former two methods, but suffers from noticeable blurring artifacts. In comparison, our proposed Laplace-Mamba effectively preserves most image details while generating a visually pleasing, haze-free result.
To further assess the haze removal capabilities of Laplace-Mamba, we present qualitative comparisons on a challenging hazy image from the LMHaze dataset in Fig. \ref{fig:fig7}. As shown, AOD-Net fails to effectively remove haze, resulting in significant residual haze and poor detail recovery. GridDehazeNet offers improved detail restoration compared to AOD-Net, but produces an overly bright and unnatural appearance. DehazeFormer achieves better detail preservation than the previous two methods but suffers from noticeable blurring artifacts. In contrast, our proposed Laplace-Mamba demonstrates superior performance, achieving a visually pleasing, haze-free image with enhanced detail preservation and natural color balance.

\begin{figure*}[ht] 
  \centering
	\includegraphics[width=0.95\linewidth]{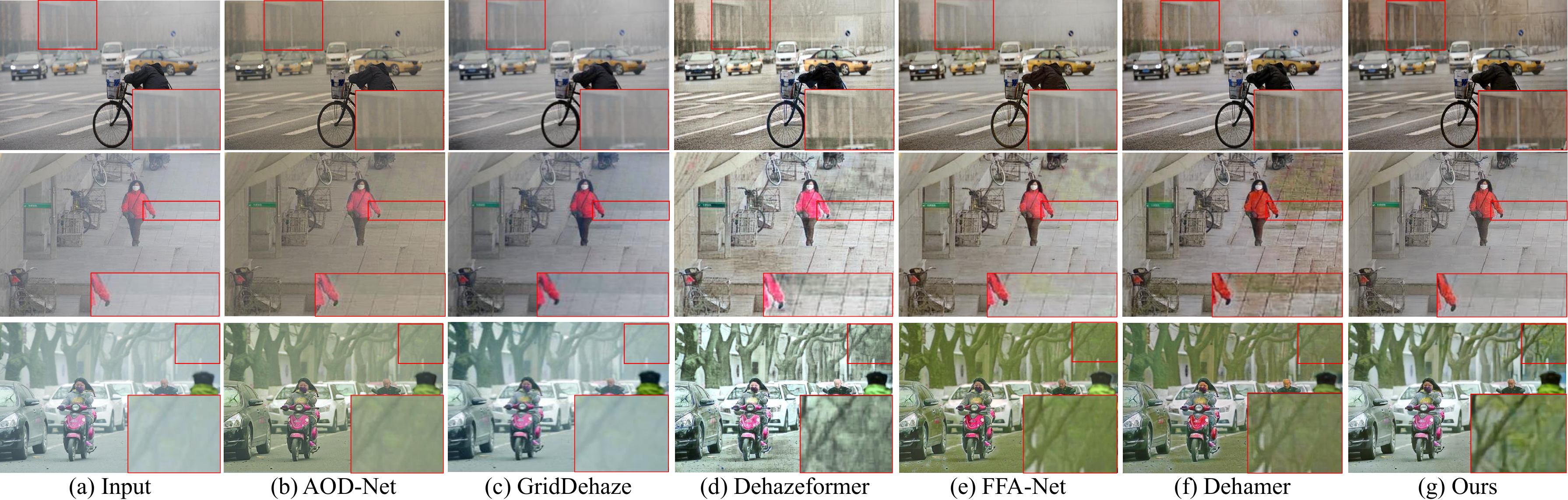}
	\caption{Qualitative comparisons on real-world hazy images, where all models are trained on the O-Haze \cite{ancuti2018haze} dataset. As observed, Laplace-Mamba produces visually superior images with more natural colors and fewer residual haze artifacts.}
	\label{fig:fig8}
\end{figure*}

\textbf{Results on Real-World Hazy Images.} To assess the dehazing performance in real-world conditions, we evaluate Laplace-Mamba on a standard real-world hazy image dataset (\textit{i.e.}, RTTS~\cite{li2018benchmarking}). As shown in Fig. \ref{fig:fig8}, for real-world hazy images, AOD-Net \cite{li2017aod}, FFA-Net \cite{qin2020feature}, and Dehamer \cite{guo2022dehamer} fail to fully eliminate the haze. While DehazeFormer \cite{song2023dehazeformer} produces relatively clear results, it often suffers from image darkening and localized blurring. Griddehazenet \cite{liu2019griddehaze} performs reasonably well in terms of haze removal but exhibits clear deficiencies in global haze suppression. In contrast, our Laplace-Mamba achieves significantly clearer dehazing, preserving image details without introducing any color distortion.

% For quantitative comparison, two well-known no-reference (BRISQUE \cite{mittal2012no} and NIQE\cite{mittal2012making}) and a reduced-reference ($\sigma$ \cite{hautiere2008blind}) image quality assessment indicators are adopted for evaluations. All these metrics are evaluated on 2,000 real-world hazy images randomly selected from the URHI dataset. Experimental results are tabulated in Table \ref{Tab:3}. The BRISQUE and NIQE metrics are used to assess the overall quality of the restored images, with lower values indicating better results. As observed, USCFormer achieves first and second place in these two indicators, respectively, showing that the haze-free images restored by our method are of high quality. $\sigma$ is used to evaluate the color restoration performance of dehazing algorithms, and our USCFormer wins first place again, indicating that USCFormer can produce haze-free images with less color distortion and fewer artifacts. Overall, USCFormer wins two of the three metrics, fully validating the superiority of our method on real-world dehazing tasks.

\subsection{Ablation Studies}
\textbf{Effectiveness of Different Modules.} 
%Through a comprehensive ablation study, we systematically validated the efficacy of three core components of our architecture: the LSRB, the HDEB, and the Multi-Domain Fusion Module (MDFM). As shown in Table~\ref{Tab:4}, we consider six variants: (1) variant 1, where replacing both the LSRB and the HDEB with simple residual blocks; (2)  replacing the MDFM with a generic concatenation operation and replacing the LSRB with a residual block; (3) replacing the LSRB with a residual block only; (4) replacing the MDFM with a standard concatenation and the HDEB with a residual block; (5) replacing the MDFM with a standard concatenation only, and (6) replacing the HDEB with a residual block only. 
We perform a comprehensive ablation study to systematically validate the effectiveness of three core components of the Laplace-Mamba architecture: the Low-frequency Structure Restoration Block (LSRB), the High-frequency Detail Enhancement Block (HDEB), and the Multi-Domain Fusion Module (MDFM). As shown in Table~\ref{Tab:4}, we evaluate six variants of the proposed method: (1) replacing both LSRB and HDEB with simple residual blocks; (2) replacing MDFM with a generic concatenation operation and LSRB with a residual block; (3) replacing LSRB with a residual block only; (4) replacing MDFM with a standard concatenation and HDEB with a residual block; (5) replacing MDFM with a standard concatenation only; and (6) replacing HDEB with a residual block only.

%The key role of the LSRB in information processing is evident. When the LSRB is replaced with a residual block, the PSNR metric decreases significantly, by approximately 1 dB, demonstrating the superiority of the structured State Space Model (SSM) for global feature modeling. For the HDEB, replacing it with a single residual block and omitting the local feature localization process causes fluctuations in PSNR of up to 2 dB, indicating that the HDEB effectively facilitates local detail restoration. Regarding the MDFM, we observe that using a simple concatenation leads to a lack of global information integration, resulting in a PSNR drop of about 1 dB. This further confirms that our MDFM better extracts and fuses global information from multiple domains.  
%The visual corroboration in Fig. \ref{fig:fig8} demonstrates that, compared to the variant with the missing component, our model can reconstruct sharper edges and more coherent textures. The complete architecture achieves state-of-the-art performance (35.70 dB PSNR, 0.991 SSIM) through well-designed frequency interactions, validating our hypothesis of complementary space-frequency processing. These systematic experiments ultimately prove the necessity of each proposed module and its collective operational mechanisms.

As shown in Table~\ref{Tab:4}, replacing the LSRB with a standard residual block leads to a notable PSNR degradation of approximately 1 dB, highlighting the advantages of the structured State Space Model (SSM) in capturing global feature representations and confirming the pivotal role of the LSRB in information processing within our architecture. Similarly, substituting the HDEB with a single residual block disrupts local feature localization, resulting in PSNR fluctuations of up to 2 dB, thereby underscoring the importance of the HDEB in preserving fine local details. In the case of the MDFM, replacing it with a simple concatenation mechanism weakens global feature integration, causing a PSNR drop of around 1 dB. These findings consistently validate that the MDFM effectively aggregates global information across multiple domains.

\begin{table}[t]
	\centering
	\footnotesize
	\caption{Ablation study of the proposed Laplace-Mamba model on the Haze4K validation set, showcasing the impact of different architectural variants on restoration performance}
	\label{Tab:4}
	\begin{threeparttable}
% 		\normalsize
		\centering
			\begin{tabular}{cccccc}
				\toprule
		    	\multirow{1}{*}{Variants}&
				\multirow{1}{*}{LSRB}&
				\multirow{1}{*}{HDEB}&
				\multirow{1}{*}{MDFM}&
				\multirow{1}{*}{PSNR$\uparrow$}&
				\multirow{1}{*}{SSIM$\uparrow$}\cr
                \hline
                $V_1$ & \ding{55} & \ding{55} &\checkmark  &34.02 &0.988 \cr
			$V_2$ &\ding{55}  & \checkmark &\ding{55}  &31.32 &0.982  \cr
                $V_3$ &\ding{55}  &\checkmark  & \checkmark  & 34.40 &0.988 \cr
                $V_4$ & \checkmark & \ding{55} &\ding{55}  & 31.91 &0.981  \cr
                $V_5$ &\checkmark  &\checkmark  &\ding{55}   &34.66  &0.989 \cr 
                $V_6$ &\checkmark  &\ding{55}  & \checkmark  &32.11  &0.984 \cr
			Ours & \checkmark  &\checkmark  & \checkmark  & \textbf{35.70} & \textbf{0.991}  \cr
			\hline
			\end{tabular}
		%}
	\end{threeparttable}
\end{table}

\begin{figure}[t] \centering
 	\includegraphics[width=\linewidth]{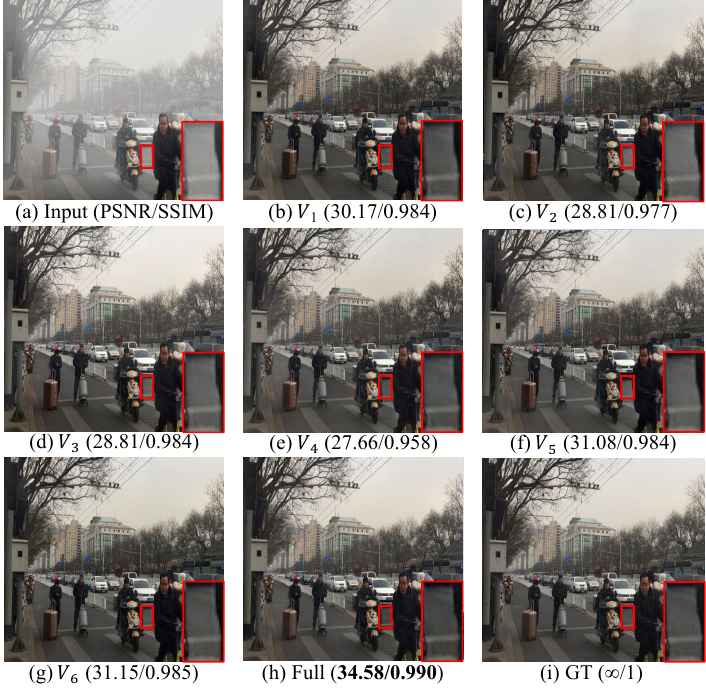}
 	\caption{Visual results of ablation studies under dense haze conditions (zoom in to see the noticeable differences).}
 	\label{fig:fig9}
\end{figure}

Qualitative comparisons in Fig. \ref{fig:fig9} further corroborate the quantitative results, where the complete Laplace-Mamba model consistently reconstructs sharper edges and maintains more coherent textures than its ablated counterparts. Equipped with all three proposed components, the full model achieves state-of-the-art performance, attaining 35.70 dB in PSNR and 0.991 in SSIM. This superior performance is attributed to the effective interactions between spatial and frequency domains, enabling comprehensive feature extraction and integration. These findings collectively demonstrate the indispensable roles of each proposed component in enhancing the dehazing capability of our method.

\textbf{Effectiveness of Different Network Configurations.} To determine the optimal network configuration, we conduct a comprehensive architectural analysis by varying the depths of the LSRB and HDEB modules. As illustrated in Table~\ref{Tab:3}, there is a clear correlation between network depth and restoration performance, measured by PSNR and SSIM. Although increasing the depth of either LSRB or HDEB leads to a higher parameter count, this does not consistently yield better results. Specifically, adding more LSRB modules in the second and third layers causes a notable drop in PSNR, while excessive HDEB stacking also degrades performance, suggesting a risk of overfitting. Through this configuration analysis, $S_1$ emerges as the optimal design, achieving a PSNR of 35.70 dB with only 9.96 M parameters. This configuration provides the optimal balance between performance and efficiency, achieving high-quality restoration with minimal computational overhead. Based on these insights, we adopt the optimized architecture to ensure the robustness and effectiveness of Laplace-Mamba’s dehazing capability.

\begin{table}[t]
	\centering
	%\scriptsize
	\caption{Ablation study on the impact of varying configurations of LSRB and HDEB modules on model performance}%, where K represents the number of LFRB and HDEB modules in each stage.}
	\label{Tab:3}
	\begin{threeparttable}
% 		\normalsize
        % \footnotesize
		\centering
        \renewcommand\arraystretch{1.2}
		\setlength{\tabcolsep}{4.0pt}{
			\begin{tabular}{cccccc}
				\hline
				% \multirow{2}{*}{Experiment}&
				% \multirow{1}{*}{LSRB}&
				% \multirow{1}{*}{HDEB}&
				% \multirow{2}{*}{PSNR$\uparrow$}&
				% \multirow{2}{*}{SSIM$\uparrow$}&
    %             \multirow{2}{*}{Parameters$\downarrow$}\cr
    %             \multirow{2}{*}{}&
				% \multirow{1}{*}{[K,K,K,K,K,K,K]}&
				% \multirow{1}{*}{[K,K,K,K,K,K,K]}&
				% \multirow{2}{*}{}&
				% \multirow{2}{*}{}&
    %             \multirow{2}{*}{}\cr
                Settings & LSRB & HDEB & PSNR$\uparrow$& SSIM$\uparrow$ & Parameters$\downarrow$\cr
				\hline
                % Exp.1(Ours) & [1,1,2,4] & [1,1,1,2] & \textbf{35.70} & 0.9907 & \textbf{9.96}\cr
                % Exp.2 & [1,1,2,4] & [1,1,2,2] & 33.41 & 0.9861 & 10.32\cr
                % Exp.3 & [1,1,4,4] & [1,1,1,2] & 35.39 & \textbf{0.9909} &10.16\cr
                % Exp.4 & [1,1,4,4] & [1,1,2,2] & 35.12 & 0.9900 &  10.45\cr
                % Exp.5 & [1,2,4,4] & [1,1,1,2] & 35.00 & 0.9904 & 10.18\cr
                % Exp.6 & [1,2,4,4] & [1,1,2,2] & 34.55 & 0.9890 & 10.47\cr
                $S_1$ & [1,1,2,4,2,1,1] & [1,1,1,2,1,1,1] & \textbf{35.70} & \textbf{0.991} & \textbf{9.96 M}\cr
                $S_2$ & [1,1,2,4,2,1,1] & [1,1,2,2,2,1,1] & 33.41 & 0.986 & 10.32 M\cr
                $S_3$ & [1,1,4,4,4,1,1] & [1,1,1,2,1,1,1] & 35.39 & \textbf{0.991} &10.16 M\cr
                $S_4$ & [1,1,4,4,4,1,1] & [1,1,2,2,2,1,1] & 35.12 & 0.990 &  10.45 M\cr
                $S_5$ & [1,2,4,4,4,2,1] & [1,1,1,2,1,1,1] & 35.00 & 0.990 & 10.18 M\cr
                $S_6$ & [1,2,4,4,4,2,1] & [1,1,2,2,2,1,1] & 34.55 & 0.989 & 10.47 M\cr
			\hline
			\end{tabular}
		}
	\end{threeparttable}
\end{table}

\subsection{Efficiency Analysis} 
To evaluate the efficiency of our Laplace-Mamba, we perform runtime assessments across 9 representative image dehazing methods. The experimental results in Table~\ref{tab5} reveal that Laplace-Mamba achieves substantial computational efficiency, processing a 400$\times$400 hazy image with only 68.90 GFLOPs and 0.189 \text{s}, while maintaining superior restoration quality. Despite the typically higher computational demands of hybrid Mamba-CNN models, which offer improved restoration performance, our implementation remains highly efficient due to three key designs: (1) parallelizable operations within the structured SSM in the frequency domain; (2) memory-efficient multi-scale feature aggregation, which enables compact yet expressive representations; and (3) a Laplace-based downsampling strategy for frequency decomposition, effectively minimizing computational cost without compromising performance. Moreover, Laplace-Mamba outperforms both Mamba-based and Transformer-based methods in terms of FLOP efficiency and inference speed. These results highlight the strong potential of spatial state-space modeling for achieving efficient and practical image dehazing.

\begin{table}[t]
\centering
	\caption{Average runtime and GFLOPS of different dehazing approaches tested on the Haze4K \cite{song2021haze4k} dataset}
	\label{tab5}
	\begin{threeparttable}
     % \footnotesize
	\centering
 		\setlength{\tabcolsep}{2pt}{
	\begin{tabular}{lcc}
			\toprule
			\multirow{1}{*}{Method}&
                \multirow{1}{*}{Runtime (\text{s}) $\downarrow$}& 
                \multirow{1}{*}{FLOPS (G) $\downarrow$} \cr
			\midrule
			% DehazeNet \cite{cai2016dehazenet}   & 0.035 & 0.58 \cr     
			AOD-Net \cite{li2017aod} &  0.011 & 0.12\cr 
			GridDehazeNet \cite{liu2019griddehaze}  &0.135 & 21.55 \cr
			Dehazeformer \cite{dong2020msbdn} & 0.280 & 277.02\cr
			FFA-Net \cite{qin2020feature} & 0.389 & 287.82 \cr
			Dehamer \cite{guo2022dehamer}  & 0.055 & 59.31 \cr
            MambaIR \cite{guo2025mambair} & 3.415 &83.33 \cr
            LMHaze \cite{zhang2022lmhaze} & 0.761 &504.84 \cr
			DEA-Net \cite{bai2024dea} & 0.037 & 32.23 \cr
            ConvIR \cite{cui2024revitalizing}  & 0.221 & 129.34 \cr
                \midrule
			Laplace-Mamba (Ours) & 0.189  & 68.90  \cr
			\bottomrule
	\end{tabular}
 		}
	\end{threeparttable}
\end{table}

% To assess the efficiency of our Laplace-Mamba, we conduct runtime evaluations across 10 representative image dehazing methods. As reported in Table~\ref{tab5}, our method demonstrates outstanding computational performance, requiring only 68.9 GFLOPs and 0.112 seconds to process a single 400$\times$400 hazy image, despite its hybrid architecture combining Mamba and CNN components. Evaluated on standardized hardware (Intel i9-12900KF CPU and NVIDIA RTX 3090 GPU), our framework achieves the second-fastest inference time among the compared approaches. This efficiency is attributed to three synergistic architectural optimizations: (1) parallelizable operations in the structured state-space model (SSM) within the frequency domain; (2) memory-efficient multi-scale feature aggregation, enabling compact yet expressive representations; and (3) a Laplace-based downsampling frequency division strategy, which minimizes computational cost without sacrificing performance. Moreover, Laplace-Mamba also achieves superior FLOPs-to-throughput efficiency and faster inference speed relative to both CNN- and Transformer-based methods. These results underscore the strong potential of spatial state-space modeling for efficient and practical image dehazing.

\section{Conclusion}
% In this work, we present Laplace-Mamba, a novel framework that seamlessly integrates a Laplace frequency prior with the Mamba-CNN architecture for efficient image dehazing. Laplace-Mamba introduces three key innovations: First, the input image is decomposed into low- and high-frequency components using prior knowledge in the frequency domain. The low-frequency components are processed by Mamba for effective global context modeling, while the high-frequency components are restored by CNN to capture fine-grained local details. This complementary design ensures accurate restoration of both global and local features. Second, the efficiency of the proposed method is enhanced by the Laplace decomposition strategy, where low-frequency components are downsampled, thereby reducing the input complexity for subsequent processing steps. Crucially, this decomposition is information-lossless, preserving full image fidelity during reconstruction. Third, to mitigate the limited representation in low-frequency components, we introduce a multi-domain fusion module. This module adaptively combines global features in both the spatial and frequency domains, significantly improving dehazing performance. Thus, leveraging the Laplace-Mamba architecture achieves exceptional results across three mainstream benchmarks: Haze4K, LMHaze, and O-Haze. The unified design outperforms state-of-the-art image dehazing methods in terms of both quantitative metrics and visual quality.

In this work, we introduce Laplace-Mamba, a novel image dehazing framework that integrates the Laplace frequency prior with a Mamba-CNN hybrid architecture through three key innovations: (1) a Frequency-Domain Collaborative Module that models low-frequency components using Mamba for global context and high-frequency components using CNNs for local detail restoration; (2) an efficient Laplace-Frequency Transform Module that decomposes low-frequency features to reduce computational complexity while preserving information fidelity; and (3) a multi-domain fusion module that adaptively integrates spatial and frequency-domain features to enhance feature representations. Extensive experiments on the Haze4K, LMHaze, and O-Haze benchmarks demonstrate that our unified approach achieves state-of-the-art performance in both quantitative metrics and visual fidelity, setting a new benchmark for efficient and effective image dehazing.

% \begin{thebibliography}{ref}
% \bibliographystyle{IEEEtran}
\bibliographystyle{IEEEtran}
\bibliography{reference}

\begin{thebibliography}{10}
\providecommand{\url}[1]{#1}
\csname url@samestyle\endcsname
\providecommand{\newblock}{\relax}
\providecommand{\bibinfo}[2]{#2}
\providecommand{\BIBentrySTDinterwordspacing}{\spaceskip=0pt\relax}
\providecommand{\BIBentryALTinterwordstretchfactor}{4}
\providecommand{\BIBentryALTinterwordspacing}{\spaceskip=\fontdimen2\font plus
\BIBentryALTinterwordstretchfactor\fontdimen3\font minus \fontdimen4\font\relax}
\providecommand{\BIBforeignlanguage}[2]{{%
\expandafter\ifx\csname l@#1\endcsname\relax
\typeout{** WARNING: IEEEtran.bst: No hyphenation pattern has been}%
\typeout{** loaded for the language `#1'. Using the pattern for}%
\typeout{** the default language instead.}%
\else
\language=\csname l@#1\endcsname
\fi
#2}}
\providecommand{\BIBdecl}{\relax}
\BIBdecl

\bibitem{he2010single}
K.~He, J.~Sun, and X.~Tang, ``Single image haze removal using dark channel prior,'' \emph{IEEE Trans. Pattern Anal. Mach. Intell.}, vol.~33, no.~12, pp. 2341--2353, 2010.

\bibitem{cai2016dehazenet}
B.~Cai, X.~Xu, K.~Jia, C.~Qing, and D.~Tao, ``Dehazenet: An end-to-end system for single image haze removal,'' \emph{IEEE Trans. Image Process.}, vol.~25, no.~11, pp. 5187--5198, 2016.

\bibitem{meng2013efficient}
G.~Meng, Y.~Wang, J.~Duan, S.~Xiang, and C.~Pan, ``Efficient image dehazing with boundary constraint and contextual regularization,'' in \emph{in Proc. IEEE Int. Conf. Comput. Vis. (ICCV)}, 2013, pp. 617--624.

\bibitem{zhu2015fast}
Q.~Zhu, J.~Mai, and L.~Shao, ``A fast single image haze removal algorithm using color attenuation prior,'' \emph{IEEE Trans. Image Process.}, vol.~24, no.~11, pp. 3522--3533, 2015.

\bibitem{berman2016non}
D.~Berman, T.~Treibitz, and S.~Avidan, ``Non-local image dehazing,'' in \emph{in Proc. IEEE/CVF Conf. Comput. Vis. Pattern Recognit. (CVPR)}, 2016, pp. 1674--1682.

\bibitem{song2021haze4k}
Y.~Song, Z.~He, H.~Qian, and X.~Du, ``Haze4k: A dehazing benchmark with 4k resolution hazy and haze-free images,'' in \emph{in Proc. IEEE Int. Conf. Comput. Vis. (ICCV)}, 2021, pp. 1234--1243.

\bibitem{zhang2018dcpdn}
H.~Zhang and V.~M. Patel, ``Densely connected pyramid dehazing network,'' in \emph{in Proc. IEEE/CVF Conf. Comput. Vis. Pattern Recognit. (CVPR)}, 2018, pp. 3194--3203.

\bibitem{li2018dehaze_cgan}
R.~Li, J.~Pan, Z.~Li, and J.~Tang, ``Single image dehazing via conditional generative adversarial network,'' in \emph{in Proc. IEEE/CVF Conf. Comput. Vis. Pattern Recognit. (CVPR)}.\hskip 1em plus 0.5em minus 0.4em\relax IEEE, 2018, pp. 8202--8211.

\bibitem{engin2018cycle_dehaze}
D.~Engin, A.~Gen{\c{c}}, and H.~K. Ekenel, ``Cycle-dehaze: Enhanced cyclegan for single image dehazing,'' \emph{arXiv preprint arXiv}, 2018.

\bibitem{wang2019aagan}
W.~Wang, A.~Wang, Q.~Ai, C.~Liu, and J.~Liu, ``Aagan: Enhanced single image dehazing with attention-to-attention generative adversarial network,'' \emph{IEEE Access}, vol.~7, pp. 173\,485--173\,498, 2019.

\bibitem{valanarasu2021transweather}
J.~M.~J. Valanarasu, V.~A. Sindagi, and V.~M. Patel, ``Transweather: Transformer-based restoration of images degraded by adverse weather conditions,'' in \emph{in Proc. IEEE Int. Conf. Comput. Vis. (ICCV)}.\hskip 1em plus 0.5em minus 0.4em\relax IEEE, 2021, pp. 2353--2363.

\bibitem{gu2023mamba}
A.~Gu and T.~Dao, ``Mamba: Linear-time sequence modeling with selective state spaces,'' \emph{arXiv preprint arXiv}, 2023.

\bibitem{vim}
L.~Zhu, B.~Liao, Q.~Zhang, X.~Wang, W.~Liu, and X.~Wang, ``Vision mamba: Efficient visual representation learning with bidirectional state space model,'' \emph{arXiv preprint arXiv}, 2024.

\bibitem{berman2018single}
D.~Berman, S.~Avidan \emph{et~al.}, ``Single image dehazing using haze-lines,'' \emph{IEEE Trans. Pattern Anal. Mach. Intell. (TPAMI)}, vol.~41, no.~3, pp. 720--734, 2018.

\bibitem{zamir2021multi}
S.~W. Zamir, A.~Arora, S.~Khan, M.~Hayat, F.~S. Khan, M.-H. Yang, and L.~Shao, ``Multi-stage progressive image restoration,'' in \emph{in Proc. IEEE/CVF Conf. Comput. Vis. Pattern Recognit. (CVPR)}, 2021, pp. 14\,821--14\,831.

\bibitem{cui2024omni}
Y.~Cui, W.~Ren, and A.~Knoll, ``Omni-kernel network for image restoration,'' in \emph{in Proc. AAAI Conf. Artif. Intell. (AAAI)}, vol.~38, no.~2, 2024, pp. 1426--1434.

\bibitem{li2018single}
B.~Li, X.~Peng, Z.~Wang, J.~Xu, D.~Feng, and W.~Zeng, ``Single image dehazing via conditional generative adversarial network,'' in \emph{in Proc. IEEE/CVF Conf. Comput. Vis. Pattern Recognit. (CVPR)}, 2018, pp. 8202--8211.

\bibitem{engin2018cycle}
D.~Engin, A.~Genç, and H.~Kemal~Ekenel, ``Cycle-dehaze: Enhanced cyclegan for single image dehazing,'' in \emph{in Proc. IEEE/CVF Conf. Comput. Vis. Pattern Recognit. workshops (CVPRW)}, 2018, pp. 825--833.

\bibitem{zhang2020nighttime}
J.~Zhang, Y.~Li, R.~H. Ng, Y.~Zheng, Z.~Li, Y.~Liu, Y.~Zheng, and R.~Tan, ``Nighttime dehazing with a multi-scale adversarial network,'' \emph{IEEE Trans. Image Process.}, vol.~29, pp. 5209--5222, 2020.

\bibitem{zhao2022local}
Y.~Zhao, J.~Zhang, M.~Li, and Y.~Zhang, ``Local-global transformer for single image dehazing,'' in \emph{in Proc. ACM Int. Conf. Multimedia (ACM MM)}, 2022, pp. 1234--1243.

\bibitem{wang2022uformer}
Z.~Wang, X.~Cun, J.~Bao, and J.~Liu, ``Uformer: A general u-shaped transformer for image restoration,'' \emph{in Proc. IEEE/CVF Conf. Comput. Vis. Pattern Recognit. (CVPR)}, pp. 17\,683--17\,693, 2022.

\bibitem{liang2021swinir}
J.~Liang, J.~Cao, G.~Sun, K.~Zhang, L.~Van~Gool, and R.~Timofte, ``Swinir: Image restoration using swin transformer,'' \emph{in Proc. IEEE Int. Conf. Comput. Vis. (ICCV)}, pp. 1833--1844, 2021.

\bibitem{chen2022u2former}
Z.~Chen, Y.~Zhang, J.~Gu, L.~Kong, and X.~Yang, ``U2former: A nested u-shaped transformer for image restoration,'' in \emph{in Proc. IEEE Int. Conf. Comput. Vis. (ICCV)}, 2022, pp. 2032--2041.

\bibitem{zamir2022restormer}
S.~W. Zamir, A.~Arora, S.~Khan, M.~Hayat, F.~S. Khan, and M.-H. Yang, ``Restormer: Efficient transformer for high-resolution image restoration,'' \emph{in Proc. IEEE/CVF Conf. Comput. Vis. Pattern Recognit. (CVPR)}, pp. 5728--5739, 2022.

\bibitem{li2022mrlpfnet}
B.~Li, Y.~Liu, P.~Wang, and Y.~Zhang, ``Multi-receptive lightweight partial fourier network for image dehazing,'' \emph{IEEE Trans. Image Process.}, vol.~31, pp. 3384--3396, 2022.

\bibitem{wang2023fftformer}
Z.~Wang, X.~Cun, J.~Bao, W.~Zhou, J.~Liu, and H.~Li, ``Fftformer: Frequency-augmented transformer for image restoration,'' in \emph{in Proc. AAAI Conf. Artif. Intell. (AAAI)}, vol.~37, no.~2, 2023, pp. 2569--2577.

\bibitem{gu2021efficiently}
A.~Gu, K.~Goel, and C.~R{\'e}, ``Efficiently modeling long sequences with structured state spaces,'' \emph{arXiv preprint arXiv}, 2021.

\bibitem{smith2022simplified}
J.~T. Smith, A.~Warrington, and S.~W. Linderman, ``Simplified state space layers for sequence modeling,'' \emph{arXiv preprint arXiv}, 2022.

\bibitem{zhu2024vision}
L.~Zhu, B.~Liao, Q.~Zhang, X.~Wang, W.~Liu, and X.~Wang, ``Vision mamba: Efficient visual representation learning with bidirectional state space model,'' \emph{arXiv preprint arXiv}, 2024.

\bibitem{guo2025mambair}
H.~Guo, J.~Li, T.~Dai, Z.~Ouyang, X.~Ren, and S.-T. Xia, ``Mambair: A simple baseline for image restoration with state-space model,'' in \emph{in Proc. Eur. Conf. Comput. Vis. (ECCV)}.\hskip 1em plus 0.5em minus 0.4em\relax Springer, 2024, pp. 222--241.

\bibitem{wang2024wave}
W.~Zou, H.~Gao, W.~Yang, and T.~Liu, ``Wave-mamba: Wavelet state space model for ultra-high-definition low-light image enhancement,'' in \emph{in Proc. ACM Int. Conf. Multimedia (ACM MM)}, 2024, pp. 1534--1543.

\bibitem{gao2024learning}
H.~Gao and D.~Dang, ``Learning enriched features via selective state spaces model for efficient image deblurring,'' in \emph{in Proc. ACM Int. Conf. Multimedia (ACM MM)}.\hskip 1em plus 0.5em minus 0.4em\relax ACM, 2024, p. 11719.

\bibitem{qin2020feature}
X.~Qin, Z.~Wang, C.~Bai, X.~Xie, and H.~Jia, ``Feature fusion attention network for single image dehazing,'' in \emph{in Proc. AAAI Conf. Artif. Intell. (AAAI)}, vol.~34, no.~07, 2020, pp. 11\,908--11\,915.

\bibitem{zheng2023frequency}
S.~Zheng, S.~Xu, C.~Wang, and J.~Zhang, ``Frequency-aware attention network for image restoration,'' \emph{IEEE Trans. Image Process.}, vol.~32, pp. 4567--4580, 2023.

\bibitem{ren2016gated}
W.~Ren, S.~Liu, H.~Zhang, J.~Pan, X.~Cao, and M.-H. Yang, ``Gated fusion network for single image dehazing,'' in \emph{in Proc. IEEE/CVF Conf. Comput. Vis. Pattern Recognit. (CVPR)}, 2016, pp. 3253--3261.

\bibitem{xu2024hcf}
S.~Xu, S.~Zheng, W.~Xu, R.~Xu, C.~Wang, J.~Zhang, X.~Teng, A.~Li, and L.~Guo, ``Hcf-net: Hierarchical context fusion network for infrared small object detection,'' in \emph{in Proc. IEEE Int. Conf. Multimedia Expo (ICME)}.\hskip 1em plus 0.5em minus 0.4em\relax IEEE, 2024, pp. 1--6.

\bibitem{zhao2016loss}
H.~Zhao, O.~Gallo, I.~Frosio, and J.~Kautz, ``Loss functions for image restoration with neural networks,'' \emph{IEEE Trans. Comput. Imaging}, vol.~3, no.~1, pp. 47--57, 2016.

\bibitem{lim2017enhanced}
B.~Lim, S.~Son, H.~Kim, S.~Nah, and K.~Mu~Lee, ``Enhanced deep residual networks for single image super-resolution,'' in \emph{in Proc. IEEE Conf. Comput. Vis. Pattern Recognit. Workshops (CVPRW)}, 2017, pp. 136--144.

\bibitem{jiang2021focal}
L.~Jiang, B.~Dai, W.~Wu, and C.~C. Loy, ``Focal frequency loss for image reconstruction and synthesis,'' in \emph{in Proc. IEEE Int. Conf. Comput. Vis. (ICCV)}, 2021, pp. 13\,919--13\,929.

\bibitem{li2017aod}
B.~Li, X.~Peng, Z.~Wang, J.~Xu, and D.~Feng, ``Aod-net: All-in-one dehazing network,'' in \emph{in Proc. IEEE Int. Conf. Comput. Vis. (ICCV)}, 2017, pp. 4770--4778.

\bibitem{liu2019griddehaze}
X.~Liu, Y.~Ma, Z.~Shi, and J.~Chen, ``Griddehazenet: Attention-based multi-scale network for image dehazing,'' in \emph{in Proc. IEEE Int. Conf. Comput. Vis. (ICCV)}, 2019, pp. 7314--7323.

\bibitem{dong2020msbdn}
H.~Dong, J.~Pan, L.~Xiang, Z.~Hu, X.~Zhang, F.~Wang, and M.-H. Yang, ``Multi-scale boosted dehazing network with dense feature fusion,'' in \emph{in Proc. IEEE/CVF Conf. Comput. Vis. Pattern Recognit. (CVPR)}, 2020, pp. 2157--2167.

\bibitem{ullah2021dmt}
H.~Ullah, K.~Muhammad, and S.~Anwar, ``Dmt-net: Dual-branch multi-scale transformer for image dehazing,'' \emph{IEEE Trans. Circuits Syst. Video Technol.}, vol.~32, no.~5, pp. 2939--2952, 2022.

\bibitem{zhang2022lmhaze}
R.~Zhang, H.~Yang, Y.~Yang, Y.~Fu, and L.~Pan, ``Lmhaze: Intensity-aware image dehazing with a large-scale multi-intensity real haze dataset,'' \emph{arXiv preprint arXiv}, 2024.

\bibitem{bai2024dea}
Y.~Bai, Y.~Zhang, W.~Li, and H.~Chen, ``Dea-net: Single image dehazing based on detail-enhanced convolution and content-guided attention,'' \emph{IEEE Trans. Image Process.}, vol.~33, pp. 1--15, 2024.

\bibitem{cui2024revitalizing}
Y.~Cui, W.~Ren, X.~Cao, and A.~Knoll, ``Revitalizing convolutional networks for image restoration via spatial-frequency collaboration,'' \emph{IEEE Trans. Image Process.}, vol.~33, pp. 1--15, 2024.

\bibitem{ancuti2018haze}
C.~O. Ancuti, C.~Ancuti, R.~Timofte, and C.~De~Vleeschouwer, ``O-haze: a dehazing benchmark with real hazy and haze-free outdoor images,'' in \emph{in Proc. IEEE Int. Conf. Comput. Vis. (ICCV)}, 2018, pp. 754--762.

\bibitem{kingma2014adam}
D.~P. Kingma and J.~Ba, ``Adam: A method for stochastic optimization,'' \emph{arXiv preprint arXiv}, 2014.

\bibitem{dong2015image}
C.~Dong, C.~C. Loy, K.~He, and X.~Tang, ``Image super-resolution using deep convolutional networks,'' \emph{IEEE Trans. Pattern Anal. Mach. Intell.}, vol.~38, no.~2, pp. 295--307, 2015.

\bibitem{wang2004image}
Z.~Wang, A.~C. Bovik, H.~R. Sheikh, and E.~P. Simoncelli, ``Image quality assessment: From error visibility to structural similarity,'' \emph{IEEE Trans. Image Process.}, vol.~13, no.~4, pp. 600--612, 2004.

\bibitem{guo2022dehamer}
M.~Guo, Z.~Wang, J.~Cao, X.~Zhang, Q.~Zhang, H.~Zhang, and D.~Tao, ``Image dehazing transformer with transmission-aware 3d position embedding,'' in \emph{in Proc. IEEE/CVF Conf. Comput. Vis. Pattern Recognit. (CVPR)}, 2022, pp. 5812--5822.

\bibitem{song2023dehazeformer}
H.~Song, M.~Zhang, and J.~Liu, ``Dehazeformer: Transformer-based image dehazing via high-frequency details enhancement,'' \emph{IEEE Trans. Image Process.}, vol.~32, pp. 3953--3966, 2023.

\bibitem{li2018benchmarking}
B.~Li, W.~Ren, D.~Fu, D.~Tao, D.~Feng, W.~Zeng, and Z.~Wang, ``Benchmarking single-image dehazing and beyond,'' \emph{{IEEE} Trans. Image Process.}, vol.~28, no.~1, pp. 492--505, 2018.

\end{thebibliography}

% \end{thebibliography}

% \newpage


\vspace{-5mm}

\vfill

\end{document}